# i2Nav-Robot: A Large-Scale Indoor-Outdoor Robot Dataset for Multi-Sensor Fusion Navigation

Hailiang Tang, Tisheng Zhang, *IEEE Member*, Liqiang Wang, Xin Ding, Man Yuan, Zhiyu Xiang, Jujin Chen, Yuhan Bian, Shuangyan Liu, Yuqing Wang, Guan Wang, and Xiaoji Niu, *IEEE Member*

***Abstract*—Accurate and reliable navigation is crucial for autonomous unmanned ground vehicles (UGVs). However, current UGV datasets fall short in meeting the demands for advancing navigation techniques due to limitations in sensor configuration, time synchronization, ground truth, and scenario diversity. Hence, we present i2Nav-Robot, a large-scale dataset designed for multi-sensor fusion navigation in indoor-outdoor environments. We integrate multi-modal navigation sensors, including the newest front-view and 360-degree solid-state LiDARs, 4-dimensional (4D) millimeter-wave (MMW) radar, stereo cameras, inertial measurement units (IMU), global navigation satellite system (GNSS) receivers, and wheeled odometers on an omnidirectional wheeled vehicle. Accurate timestamps are obtained through both online hardware synchronization and offline calibration for all sensors. The dataset includes ten large-scale sequences covering diverse UGV operating scenarios, such as outdoor streets and indoor parking lots, with a total length of about 17060 meters. High-rate, reliable, and fully covered ground truth, with centimeter-level positioning, is derived from post-processing integrated navigation methods using a high-grade IMU. The proposed i2Nav-Robot dataset is evaluated by 15 open-sourced multi-sensor fusion navigation methods, demonstrating its superior data quality and utility for advancing vehicular navigation research. The i2Nav-Robot dataset together with the documents can be accessed on GitHub (https://github.com/i2Nav-WHU/i2Nav-Robot).**

## I. Introduction

UNMANNED ground vehicles (UGV) are a kind of robot in contact with the ground, such as indoor buildings and outdoor roads. Nowadays, the UGV has been widely employed for logistics, cleaning, and patrol, especially with the rapid development of autonomous driving [1]. As the foundation of perception, planning, and control [2], accurate and reliable position and attitude are of great importance for autonomous UGV applications. The global navigation satellite system/inertial navigation system (GNSS/INS) integrated system [3], [4], as a classical method, has already become a basic part of UGVs. For civil applications, the intelligent UGV usually works in closed environments, such as campus, park, and residential district. In such complex environments, the GNSS signals may be interfered with, decayed, and even interrupted, and thus the navigation accuracy may be degraded [5]. Due to time-varying biases of the low-cost micro-electro-mechanical systems (MEMS) inertial measurement unit (IMU), extra information should be adopted to mitigate the INS drift in GNSS-denied scenarios.

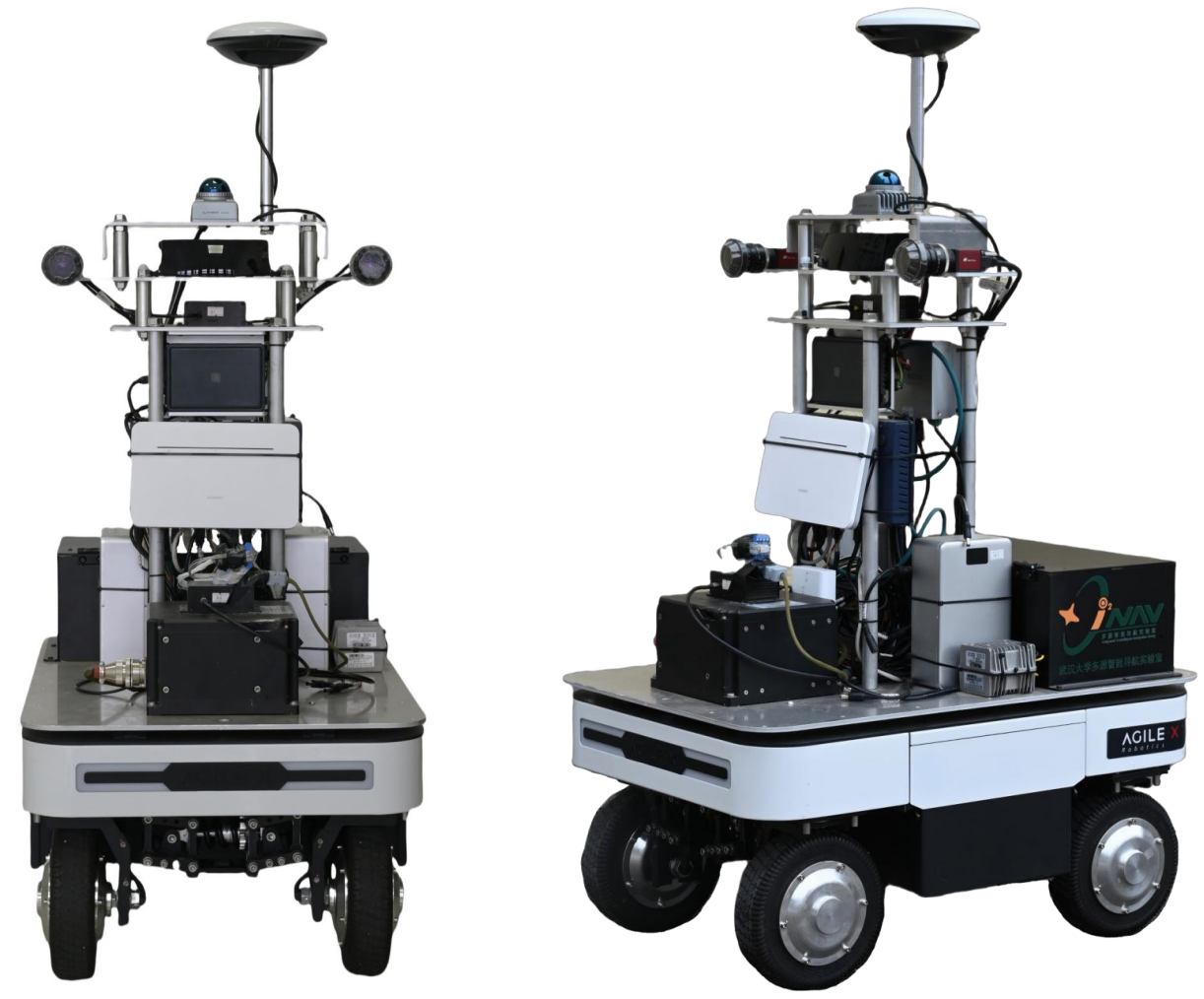

(a) Front view. (b) Oblique view.

Fig. 1. The four-wheel omnidirectional vehicle in the i2Nav-Robot dataset.

Different from unmanned aerial vehicles (UAV) [6] and unmanned surface vehicles (USV), the wheeled UGV is usually equipped with a low-cost odometer (ODO) on each wheel [7]. Hence, we can obtain precise speed measurements from the wheeled odometer (ODO). By incorporating the wheel speed and optional wheel angle, the INS drifts can be reduced notably in a GNSS/INS/ODO integration system [8], [9]. Here, we treat the wheel speed and wheel angle as all measurements of an ODO for a steering wheel. Nevertheless, the navigation accuracy still cannot be guaranteed in environments with long-term GNSS interruptions, such as in urban canyons and indoor parking lots. Thus, the LiDAR and visual simultaneous localization and mapping (SLAM) methods [10], [11], [12] can be employed to enhance the reliability and robustness. As a result, a multi-sensor fusion navigation system can be constructed, and it is a practical and reliable solution for autonomous UGV applications.

However, current UGV datasets have been unable to meet the demand of advancing navigation, especially when the newest solid-state LiDARs and 4-dimensional (4D) millimeter-wave (MMW) radars have been equipped in newly mass-produced UGVs. Table I exhibits a comparison of current

This research is funded by the National Natural Science Foundation of China (42374034 and 42504027), the Key Research and Development Program of Hubei Province (2024BAB024), and the China Postdoctoral Science Foundation under Grant Number 2025M780217. (*Corresponding author: Tisheng Zhang.*)

The authors are with the Intelligent and Integrated Navigation Group (i2Nav), GNSS Research Center, Wuhan University, Wuhan 430079, China. Hailiang Tang, Tisheng Zhang, and Xiaoji Niu, are also with the Hubei Luojia Laboratory, Wuhan 430079, China. (e-mail: zts@whu.edu.cn)

TABLE I
COMPARISON OF THE OPEN-SOURCED UGV DATASETS

| Dataset, Year | Overview | | | | Sensors | | | | | Ground truth | | | Scenarios[6] | | |
|---|---|---|---|---|---|---|---|---|---|---|---|---|---|---|---|
| | UGV[1] | Sync.[2] | All in one[3] | Large scale[4] | Vision | Front-view LiDAR | 360-degree LiDAR | 4D Radar | ODO[5] | Pose | Velocity | Rate (Hz) | IN | OUT | IO |
| NCLT, 2016 | Two-wheel Segway | Software | | ● | ● | | ● | | Pose | ● | | 100 | | ● | ● |
| Rosario, 2019 | Four-wheel Ackerman | Software | ● | | ● | | | | ● | Position Only | | 5 | | ● | |
| M2DGR, 2022 | Four-wheel Ackerman | Software | ● | | ● | | ● | | | ● | | 10~100 | ● | ● | ● |
| FusionPortableV2 (UGV), 2024 | Four-wheel Ackerman | Hardware | ● | | ● | | ● | | ● | ● | | 30 | | ● | |
| M3DGR, 2025 | Four-wheel Ackerman | Software | ● | | ● | ● | ● | | ● | Half[7] | | 5 | ● | ● | |
| **i2Nav-Robot** (ours), 2025 | Four-wheel omnidirectional | Hybrid[8] | ● | ● | ● | ● | ● | ● | ● | ● | ● | 200 | ● | ● | ● |

[1]The UGV type. [2]The time-synchronization mode for different sensors. [3]All sensors' data for each sequence is contained in a single file, typically the ROS bag file. [4]Datasets with every sequence longer than 1000 meters are defined as large-scale datasets. [5]The wheeled odometer (ODO) denotes the wheel speeds and optional wheel angles. [6]IN is indoor; OUT is outdoor; IO denotes that there are both indoor and outdoor environments in a single sequence. [7]There are ground-truth poses for indoor sequences from Mocap, while only positions for outdoor sequences from GNSS. [8]Using hardware synchronization for supported sensors, otherwise using software synchronization with offline calibrations.

state-of-the-art (SOTA) UGV datasets. According to the results in Table I, some datasets have an incomplete sensor configuration, because LiDAR, 4D radar, or ODO are not employed. In addition, most datasets employ inaccurate software synchronization to obtain timestamps of sensors, and this may result in large performance loss. Meanwhile, reliable and accurate ground-truth pose is not fully available in some datasets, exhibiting poor applicability. The employed ground-truth systems also limit the scenario diversity. In the following subsections, we will discuss the current datasets in terms of the sensor configuration, time synchronization, and ground-truth system, and present our contributions.

### *A. Sensor Configuration*

In terms of sensor configuration, a GNSS receiver, a MEMS IMU, and several ODOs have already been the standard components of commercial wheeled UGVs. This is because GNSS/INS/ODO integration can complete most navigation tasks on large-scale outdoor roads, which are the common application scenarios for a UGV. Among these sensors, the GNSS receiver is particularly important for large-scale applications, because it can provide accurate absolute positioning results. However, some datasets, such as OpenLORIS-Scene [13], Kimera-Multi [14], MCD [15], BotanicGarden [16], and Oxford Spires [17], mainly focus on SLAM methods, and they do not contain a GNSS receiver. Hence, these datasets without GNSS are not very suitable for large-scale multi-sensor navigation. Besides, the ODO is a low-cost and precise sensor, which is very useful in GNSS-denied environments. However, the M2DGR [18] dataset does not contain a single ODO.

Vision sensors or RGB cameras have been used in nearly all UGV datasets, due to the lower cost and comparable accuracy. Besides, multiple cameras are commonly equipped to increase the field of view (FOV), and the stereo camera system has become the most popular and practical configuration for navigation applications. Unlike vision sensors, the 3D LiDAR can conduct distance measurement directly, and it can achieve more robust and accurate navigation. Besides, the 3D LiDAR can also be used to build a point-cloud map, which can be further employed for map-matching-based navigation. Nevertheless, there is no LiDAR in the Rosario [19] dataset. Besides, the LiDAR sensor has been developed rapidly in the past decade, and the conventional spinning LiDAR has been gradually replaced by newly (half)-solid-state LiDARs. For example, the M3DGR [20] dataset employs a Livox AVIA front-view LiDAR and a Livox Mid360 360-degree LiDAR, which are all solid-state LiDARs. MCD [15] and BotanicGarden [16] datasets also equip a Livox LiDAR. Recently, a kind of solid-state automotive-grade LiDAR with a horizontal FOV of about 120° and a vertical FOV of about 25°, such as the RoboSense M1 and Hesai AT128, has been applied in newly autonomous vehicles and UGVs. However, such solid-state LiDARs have not yet been included in the current UGV datasets.

Like solid-state LiDARs, the 4D millimeter-wave radar has already been equipped in recent autonomous UGVs for robust perception in extreme environments. The 4D radar can be employed as a supplement to environment perception by outputting object-detection results, and it can also provide raw distance, evaluation angle, azimuth angle, and Doppler velocity [21] these 4D data. So far, the 4D radar has been contained in some vehicle or handheld datasets, such as EU Long-term [22] and NTU4DRadLM [23], but not in current UGV datasets. Hence, there is a large gap between current UGV datasets and mass-produced UGVs, in terms of the LiDAR and 4D radar.

### *B. Time Synchronization*

When heterogeneous sensors, including GNSS, MEMS IMU, camera, LiDAR, and 4D radar, are installed in the same UGV,

the time synchronization becomes a complex problem. As different sensor data is asynchronously obtained from different interfaces, such as the universal serial bus (USB) port, serial port, and Ethernet port. A naive solution is to stamp the sensor data on the host computer with the receiving time, and this is called software synchronization. Datasets like NCLT [24], Rosario [19], M2DGR [18], and M3DGR [25] all employ software synchronization. However, the software synchronization is inaccurate, and this may cause large errors and reduce the navigation reliability. In M2DGR [18], the authors claim that they achieve software synchronization of less than 10 ms. However, a time error of 10 ms will result in an angle error of 0.5° when the angular velocity is 50 °/s, which is a normal value for a wheeled UGV. For this reason, NCLT [24] and OpenLORIS-Scene [13] datasets try to calibrate the time offsets caused by software synchronization. However, the unstable time intervals cannot be calibrated, and this may introduce larger errors for navigation, as shown in Fig. 2.

The hardware synchronization using embedded triggers can achieve a microsecond-level synchronization accuracy, and it has been employed in many datasets, such as FusionPortable [26], FusionPortableV2 [27], and MARS-LVIG [28]. Meanwhile, the precision time protocol (PTP) can also be used to synchronize different sensors through an Ethernet interface, and it almost achieves the same accuracy as the trigger-based synchronization. PTP has been supported by recent LiDARs and cameras. Besides, data acquisition and time synchronization can be completed simultaneously through one single Ethernet port. As a result, the trigger-based and PTP-based hardware synchronization can be utilized together for accurate time synchronization, such as in the BotanicGarden [16] dataset.

### C. Ground-Truth System

Ground truth is another critical part of a practical dataset. For outdoor environments, GNSS real-time kinematic (RTK) with a centimeter-level accuracy can be adopted as ground-truth positions, such as in the Rosario [19] and M3DGR [20] datasets. However, the GNSS RTK usually has a very low data rate, which is not enough for a complete evaluation. Besides, the ground-truth attitudes cannot be obtained with only one GNSS antenna. Hence, the low-cost GNSS/INS integration, which can provide both positions and attitudes, is employed as the ground-truth system in FusionPortableV2 [27], but the ground-truth pose is absent in GNSS-denied environments. The NCLT [24] dataset employs both the GNSS RTK and LiDAR SLAM to generate ground-truth poses for indoor-outdoor environments, but the pose accuracy cannot be guaranteed in large-scale indoor environments.

The motion capture (Mocap) system has been a common ground-truth system for indoor environments, and the pose rate can be more than 100 Hz. For example, the famous EuRoC UAV [29] datasets, and M2DGR [18] and M3DGR [20] datasets all use the Mocap as the ground-truth system. Nevertheless, the Mocap is only suitable for an indoor laboratory or a certain region, because the working region of the Mocap is limited by the expensive high-speed camera.

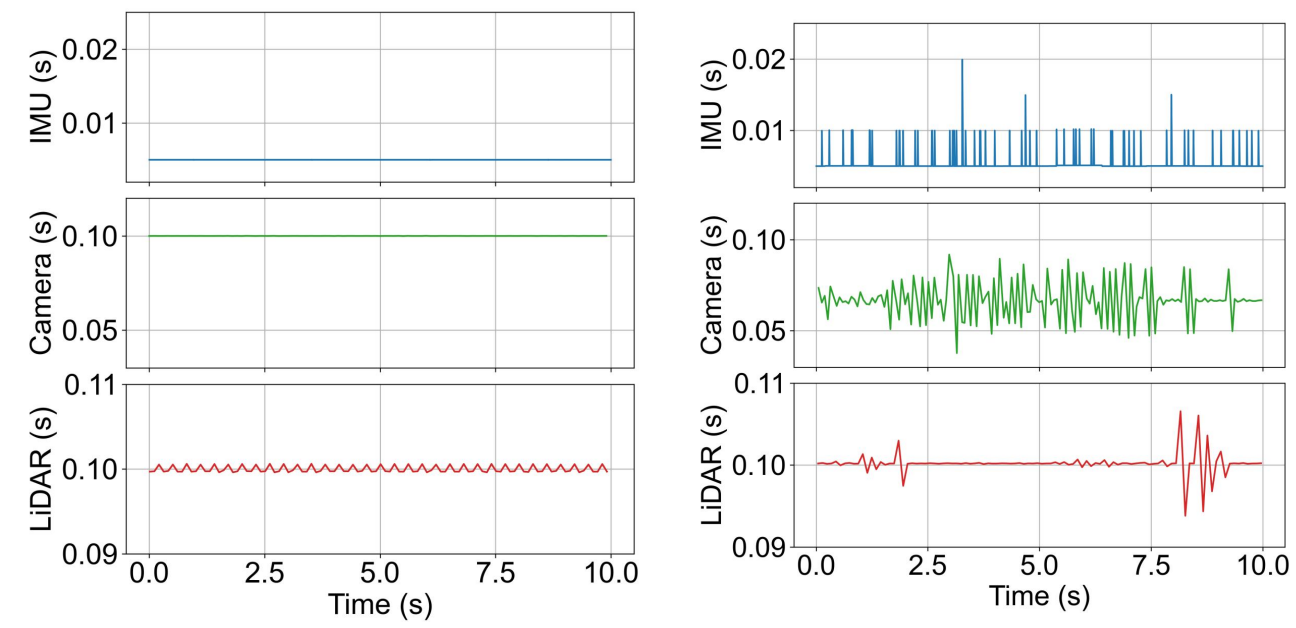

(a) Hardware synchronization. (b) Software synchronization.

Fig. 2. Comparison of the intervals of sensors' timestamps. The intervals obtained from the hardware synchronization are much steadier than that of the software synchronization, which means more accurate synchronization.

Hence, it is challenging to generate reliable ground-truth poses in large-scale indoor environments, which limits the testing environments of the current datasets. Meanwhile, most of the current datasets do not provide ground-truth velocity, which can be employed for many usages, such as accuracy evaluation and time calibration.

### D. Our Contributions

Overall, existing UGV datasets have exhibited some deficiencies in sensor configuration, time synchronization, ground truth, and scenario diversity, as shown in Table I. Specifically, the main limitations of current SOTA UGV datasets in Table I are: 1) Inaccurate software synchronization is employed in the NCLT, Rosario, M2DGR, and M3DGR datasets. 2) Modern sensors, such as LiDAR and 4D radar, and the ODO are not included in some datasets, like Rosario and M2DGR datasets. 3) Accurate ground-truth pose cannot be ensured in some sequences of Rosario and M3DGR datasets. 4) Most datasets cannot cover both indoor and outdoor scenarios, which is limited by their ground truth system. 5) Except for the NCLT dataset, other datasets in Table I are all collected in small-scale environments with trajectories length less than 1,000 meters. 6) The indoor scenes in M2DGR and M3DGR, such as the motion-capture room, are not the typical application scenarios for UGVs.

Hence, we present the i2Nav-Robot dataset with multi-modal sensors, accurate time synchronization, reliable ground truth, and abundant scenarios. Fig. 1 shows the four-wheel omnidirectional robot adopted in the i2Nav-Robot dataset. The presented i2Nav-Robot dataset is suitable for many multi-sensor fusion navigation applications, such as visual-inertial odometry (VIO), LiDAR-inertial odometry (LIO), GNSS/INS integration, and LiDAR-visual-inertial odometry. The main contributions of our work are as follows:

- Multi-modal sensors, including front-view solid-state LiDAR, 360-degree solid-state LiDAR, 4D MMW radar, stereo cameras, IMU, GNSS receivers, and wheeled ODO, are well integrated on an omnidirectional robot, enabling complete motion perception for accurate navigation.
- Both the trigger-based and PTP-based hardware synchronizations are employed to obtain microsecond-level timestamps for nearly all sensors. Besides, the fixed time offsets for the software-synchronized radar and ODO are calibrated offline using a speed-based method to achieve a millisecond-level synchronization.

● Ten large-scale sequences with a total length of more than 17000 meters and 14000 seconds are collected in diverse environments, such as outdoor streets, indoor parking lots, and indoor-outdoor buildings, which are all potential application scenarios for a commercial UGV.

● A navigation-grade IMU is adopted to provide high data rate ground-truth position, attitude, and velocity for all sequences. The post-processing GNSS/INS integration is adopted to derive centimeter-level ground truth for outdoor sequences, and the post-processing map-matching (MM)/INS integration is used for GNSS-challenging sequences.

● The proposed dataset is examined by 15 open-sourced multi-sensor navigation methods. Sufficient results demonstrate the extraordinary data quality of the i2Nav-Robot dataset. Besides, experiment results show that the sensors are well synchronized, and the provided calibration parameters are exactly accurate.

In the following sections, we first give an overview of the i2Nav-Robot dataset. In Section III, the details of the ground-truth system are presented. Then, the dataset descriptions are introduced in Section IV. Experiments and results for quantitative evaluation are discussed in Section V. Finally, we make a brief conclusion of the presented i2Nav-Robot dataset.

## II. System Overview

### A. Sensor Setup

The multi-sensor robot employed in the i2Nav-Robot dataset is built upon a four-wheel omnidirectional robot, *i.e.,* AgileX Ranger Mini3, as shown in Fig. 1 and Fig. 3. Table II exhibits the specifications of the equipped multiple sensors. A Jetson AGX Orin board from NVIDIA is employed to collect all sensor data through the Robot Operating System (ROS). The Ranger chassis can output wheel speeds and angles from the four wheels to the Orin through a Controller Area Network (CAN) bus. INS-Probe, a compact GNSS/INS integrated system developed by the i2Nav group, provides the IMU data from ADI ADIS16465 and the GNSS data from Ublox F9P. We also equipped a NovAtel OEM719 receiver to derive GNSS-RTK positioning results and GNSS raw observations. The INS-Probe and OEM719 are connected to the Orin through USB ports. The ground-truth GNSS/INS system includes a navigation-grade IMU with a gyroscope bias stability of 0.03 °/hr, which is used to generate the ground truth by integrating with the post-processing RTK or MM results. The GNSS signals for INS-Probe, OEM719, and ground-truth system are derived from a full-frequency antenna, *i.e.,* Harxon CSX601A, through a multi-port splitter.

Multi-modal perception sensors, including stereo cameras Mako-G234 from AVT, a front-view solid-state LiDAR AT128 from Hesai, a 360-degree solid-state LiDAR Mid360 from Livox, and a 4D MMW radar ARS548 from Continental, are installed on the robot using our self-designed mechanical structure. In particular, the AT128 has a horizontal FOV of 120° and a vertical FOV of 25°, and this kind of LiDAR has become a substitute for conventional 360-degree spinning LiDARs in mass-produced UGVs. Meanwhile, the ARS548 is a new 4D MMW radar, which can provide detected objects, such as pedestrians and vehicles. The ARS548 can also provide point clouds and Doppler velocity measurements, which may be particularly useful for robust navigation, especially in visual and LiDAR degraded environments. Fig. 4 depicts the multi-modal perception from camera, front-view LiDAR, 360-degree LiDAR, and 4D MMW radar. By fully using these sensors, we can achieve reliable and accurate multi-sensor fusion navigation in complex scenarios.

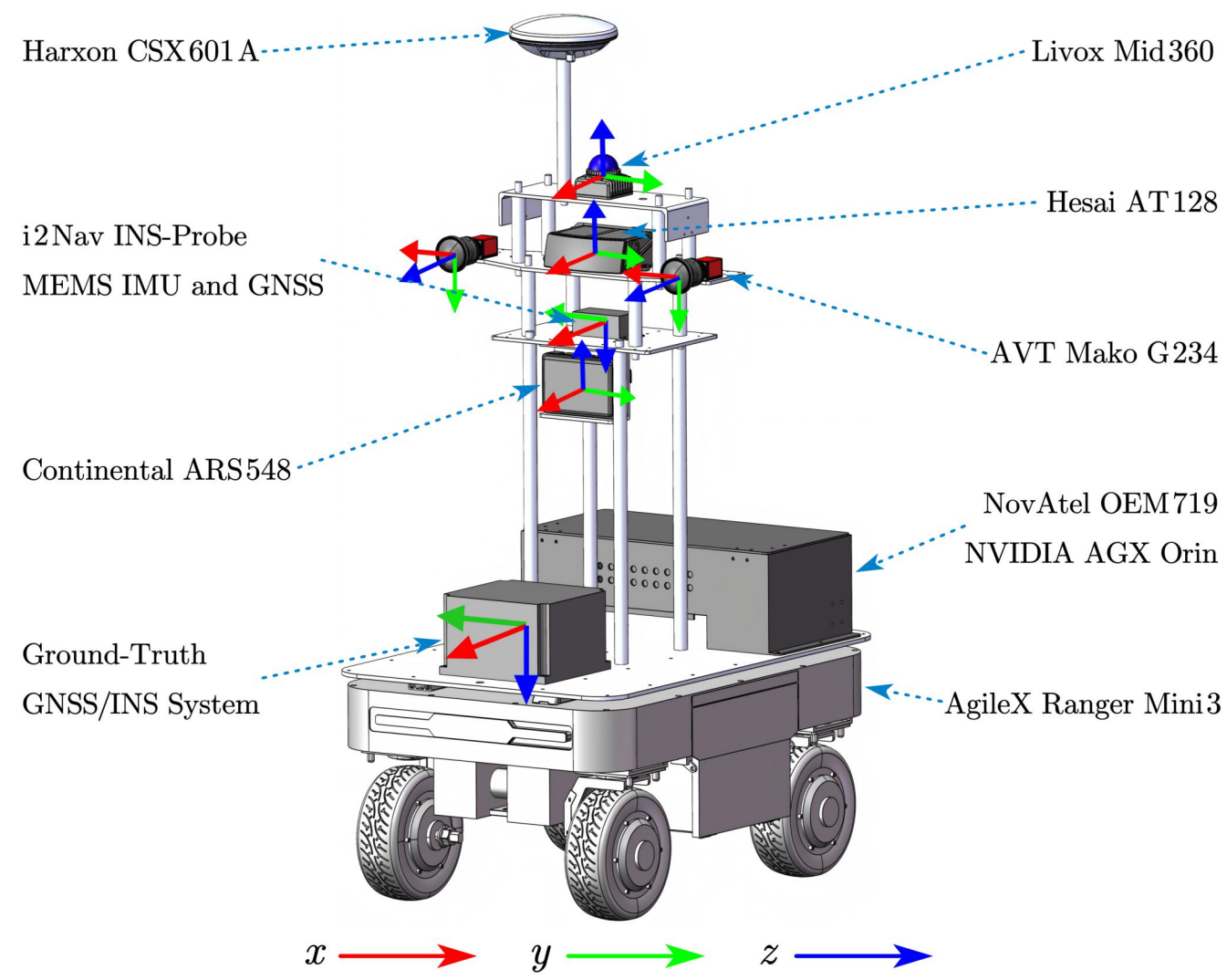

Fig. 3. The 3D model of the robot together with coordinate frames.

TABLE II
The Equipped Sensors and Specifications

| Sensor | Type | Characteristic |
|---|---|---|
| ADI ADIS16465 | IMU | In-run bias stability: 2 °/hr<br>Angular random walk: 0.15 °/√hr |
| Ublox F9P | GNSS | Multi-system supported<br>L1/L2 double frequency |
| Harxon CSX601A | GNSS antenna | Full GNSS frequency supported |
| NovAtel OEM719 | GNSS | Multi-system and multi-frequency supported |
| AVT Mako-G234 (Stereo) | Camera | Global shutter and color mode<br>Resolution: 1600*1200 |
| Livox Mid360 (Built-in ICM40609) | 360-degree LiDAR | Scanning line: 6<br>Horizontal: 360°, vertical: -7°~52°<br>Range: 0.1~40 m @ 10% reflectivity |
| TDK ICM40609 | IMU | Gyroscope noise: 4.5 mdps/√Hz<br>Accelerometer noise: 100 ug/√Hz |
| Hesai AT128 | Front-view LiDAR | Scanning line: 128<br>Horizontal: 120°, vertical: -12.5°~12.9°<br>Range: 1~180 m @ 10% reflectivity |
| Continental ARS548 | 4D Radar | Horizontal: 120°, vertical: -20°~20°<br>Range: 0.2~300 m, accuracy: ±0.15 m<br>Speed accuracy: 0.1 m/s |
| AgileX Ranger Mini3 | Wheeled odometer | Speed resolution: 0.001 m/s<br>Angle resolution: 0.001 rad |

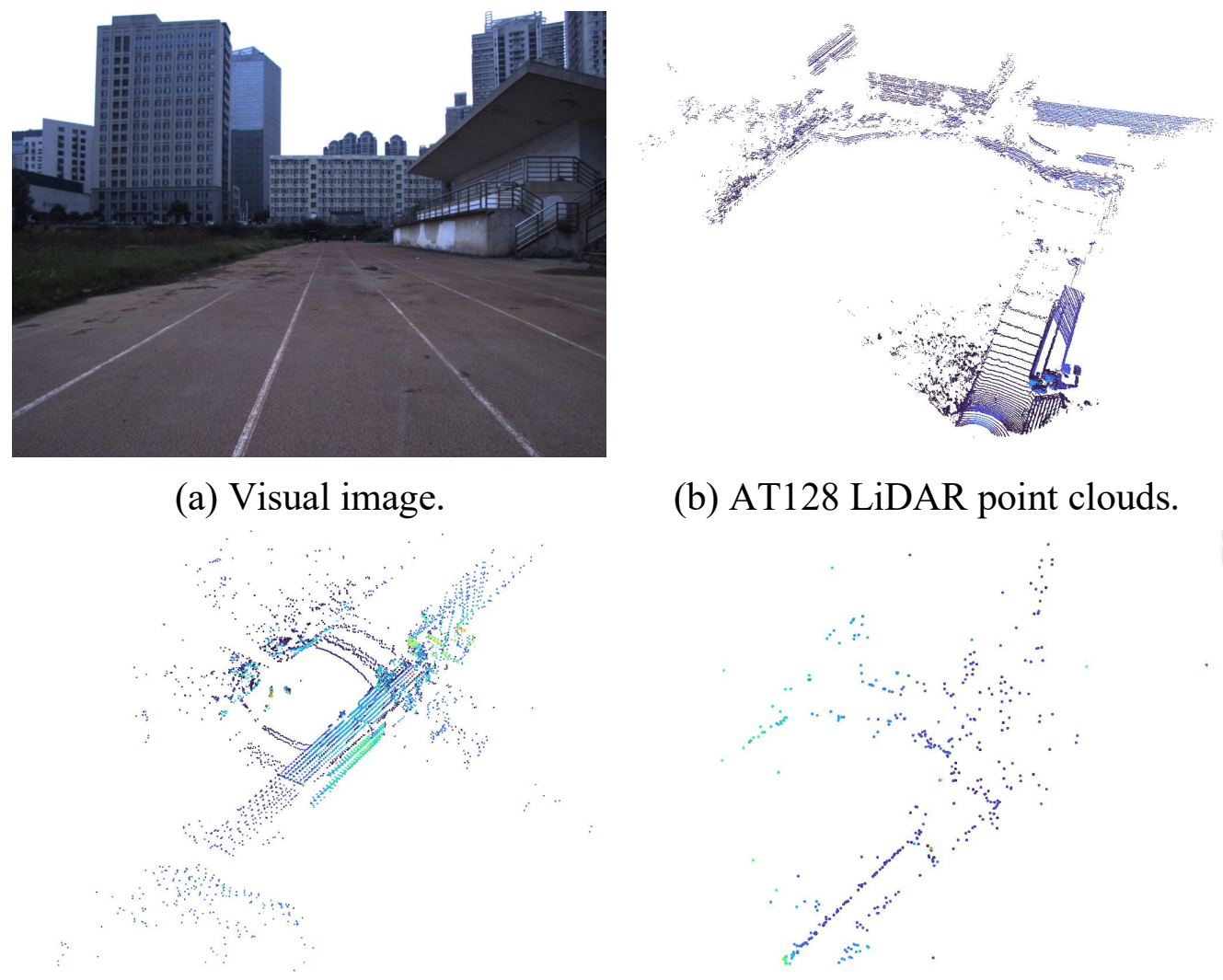

(a) Visual image. (b) AT128 LiDAR point clouds.

(c) Mid360 LiDAR point clouds. (d) ARS548 radar point clouds.

Fig. 4. Multi-modal perception from camera, front-view LiDAR (AT128), 360-degree LiDAR (Mid360), and 4D MMW radar (ARS548).

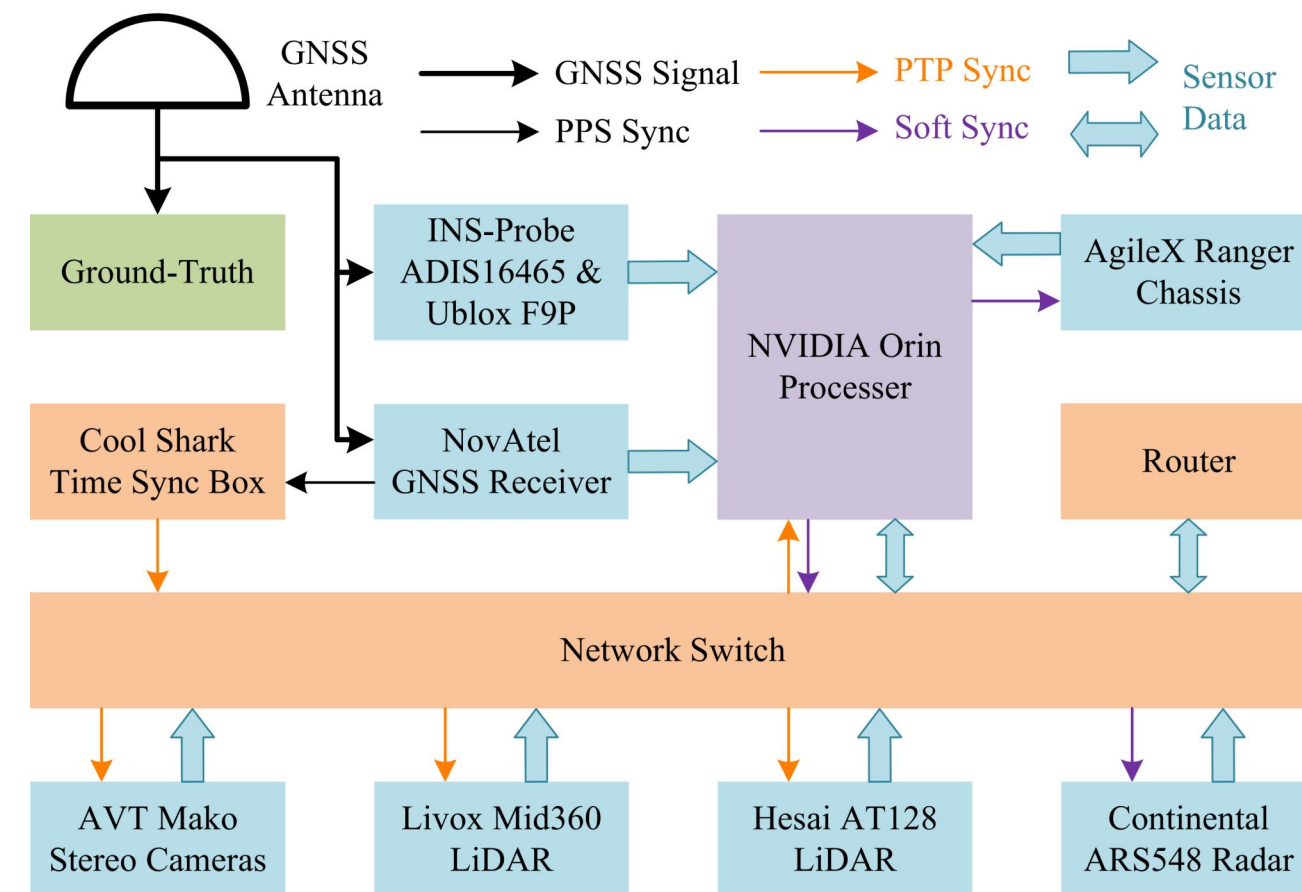

Fig. 5. Time-synchronization flow.

## B. *Time Synchronization*

Accurate time synchronizations for all sensors are achieved through both hardware synchronizations and offline calibrations. The trigger-based and PTP-based hardware synchronizations are employed to obtain microsecond-level timestamps for GNSS receivers, IMUs, LiDARs, and cameras. Besides, the fixed time offsets for the software-synchronized radar and ODO are calibrated offline using a speed-based method to achieve a millisecond-level synchronization.

Specifically, GNSS is the time source for all the systems through the pulse-per-second (PPS) signal, as shown in Fig. 5. The Cool Shark box is a module for PTP-based time synchronization. The employed PTP protocol follows the IEEE standard 1588-2008 [30]. The ground-truth system, INS-Probe, and Cool Shark box can maintain accurate clocks with their built-in crystal oscillators, even in GNSS-denied indoor environments. Besides, the ADIS16465 IMU in the INS-Probe is synchronized through hardware trigger. Meanwhile, the Cool Shark box is the main clock for PTP through a local network switch. Hence, the cameras, LiDARs, and NVIDIA Orin can be synchronized to the GNSS time using PTP. These sensors synchronized through hardware can have microsecond-level timestamps. Note that the timestamps for stereo cameras correspond to the start time of exposure, and thus the exposure durations are not considered in the timestamps.

The ARS548 radar and AgileX ODO are synchronized using the system time on Orin when receiving the data, because they do not have a hardware-synchronization interface. Nevertheless, they are also precisely synchronized, as the NVIDIA Orin has been synchronized through PTP. Besides, the fixed time offsets for radar and ODO data can be calibrated offline.

## C. *Calibration*

### 1) *Calibration of Intrinsic and Extrinsic Parameters*

The intrinsic and extrinsic parameters are important for accurate navigation. In the i2Nav-Robot dataset, the stereo cameras are modeled using the pinhole camera and radial-tangential distortion models [31]. The intrinsic and extrinsic parameters for stereo cameras are calibrated using MATLAB 2024b with an AprilTag [32] calibration board. The translation parts of the extrinsic parameters for all sensors are directly obtained from their 3D models, as exhibited in Fig. 3. We do not calibrate the rotation parts of the extrinsic parameters. Nevertheless, we provide the LiDAR-IMU and camera-IMU extrinsic parameters estimated by the open-sourced FF-LINS [33] and LE-VINS [34] which are included in the calibration file. The LiDAR-LiDAR [35] and LiDAR-camera [36] extrinsic parameters can be calculated using parameters in the calibration file. Besides, experiments will be conducted to verify the accuracy of the provided extrinsic parameters in Section V.C. These parameters are recommended as the initial values when using the presented dataset.

### 2) *Calibration of Time Offsets*

The 4D radar and ODO are synchronized through software, due to the lack of hardware-synchronization interface. Nevertheless, we employ a speed-based method to calibrate the fixed time offsets and improve the accuracy of time synchronization. For the 4D radar, we can obtain hundreds of points $\boldsymbol{p}_{R,i}$ with Doppler speeds $v_{R,i}$, where $i$ is the point index. Hence, we can estimate the radar velocity $\boldsymbol{v}_R$ by solving the following least-square problem as

$$arg \min_{\boldsymbol{v}_R} \sum_i \left\| \boldsymbol{v}_R^T \cdot \boldsymbol{u}_{R,i} - v_{R,i} \right\|^2 \tag{1}$$

where $\boldsymbol{u}_{R,i} = \boldsymbol{p}_{R,i} / \left\| \boldsymbol{p}_{R,i} \right\|$ is the direction vector of $\boldsymbol{p}_{R,i}$. The radar speed $v_{Radar}$ is the norm of the radar velocity $\boldsymbol{v}_R$.

For the ODO, the wheel speed and angle for each wheel of the four-wheel omnidirectional robot can be expressed as $v_j$ and $\alpha_j$, where $j \in [1,4]$. The ODO speed $v_{ODO}$ in the vehicle frame, derived from the four wheel speeds and angles, can be written as

$$v_x = \frac{1}{4} \sum_{j=1}^{4} v_j * \cos \alpha_j \tag{2}$$

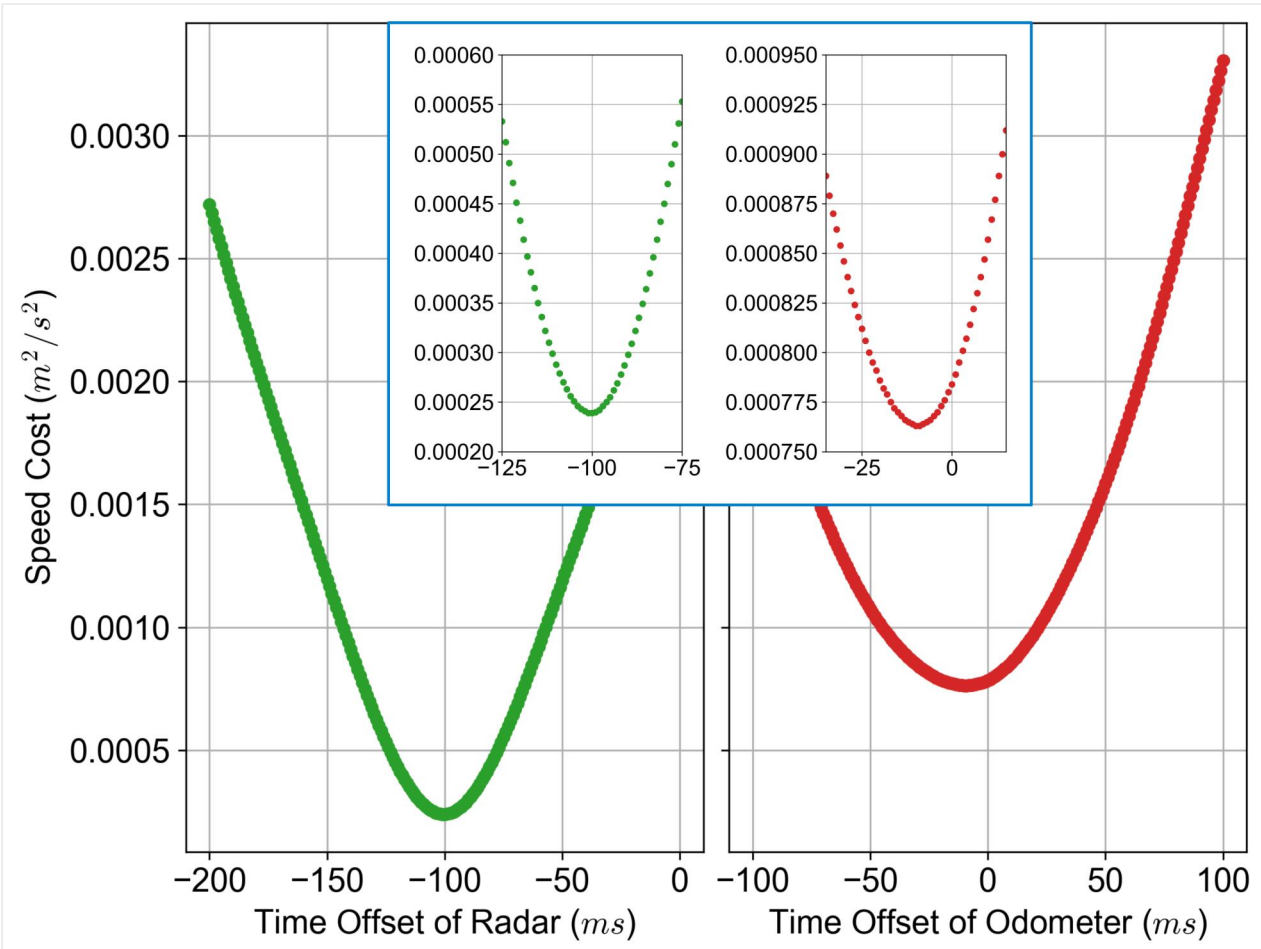

Fig. 6. Comparison speed costs of radar and ODO over different time offsets. The time offset corresponding to the minimum speed cost is the optimal estimation. The time offsets for radar and ODO in this figure are about -100 ms and -10 ms, respectively.

$$v_y = \frac{1}{4}\sum_{j=1}^{4} v_j * \sin\alpha_j \quad (3)$$

$$v_{ODO} = \sqrt{v_x^2 + v_y^2} \quad (4)$$

The $v_x$ and $v_y$ denote the forward and lateral speed, respectively. $v_y$ is almost equal to zero due to the dynamic model of the omnidirectional robot [8], but it is reserved to ensure the integrity of the formula. The calculated ODO speed series from (4) are available in our GitHub repository for users' convenience.

As a result, we can derive the time series of the radar and ODO speed from (1) and (4). The fixed time offsets can be estimated by conducting a least-square method with the ground-truth speed series. The speed series from radar, ODO, and ground truth are first interpolated into 1,000 Hz, to obtain an estimation accuracy of about 1.0 ms. Giving a time offset $\delta t$ with a resolution of 0.001 s, the index offset $M$ of the 1,000 Hz speed series can be written as $M = 1000\delta t$. Then, we can calculate the differences $\delta v_k$ between the estimated ODO or radar speed $v_{Est,k+M}$ and the ground-truth speed $v_{GT,k}$ as

$$\delta v_k = v_{Est,k+M} - v_{GT,k} \quad (5)$$

where $k$ is an index. Finally, the time offset $\delta t$ or the index offset $M$ can be obtained by minimizing the average of $\|\delta v_k\|^2$, *i.e.* the speed cost, using a speed series with $N$ indices as

$$arg\min_{M}\frac{1}{N}\sum_{k}\|\delta v_k\|^2 \quad (6)$$

Note that using a time series segment with aggressive motions will lead to more accurate time-offset estimation.

Fig. 6 compares speed costs of radar and ODO over different time offsets. The optimal estimation of the time offset can be obtained from the minimum speed cost. According to the results in Fig. 6, the fixed time offset for the radar is about -100 ms, while about -10 ms for the ODO. The negative signs denote that the radar and ODO data are delayed. As depicted in Fig. 7, the radar and ODO speeds are synchronized to the ground truth speed after compensating for the time offsets. Hence, the radar and ODO can also achieve a synchronization accuracy of 1.0 ms, which is much better than the uncalibrated software synchronization. The time-offset parameters are included in the provided calibration file.

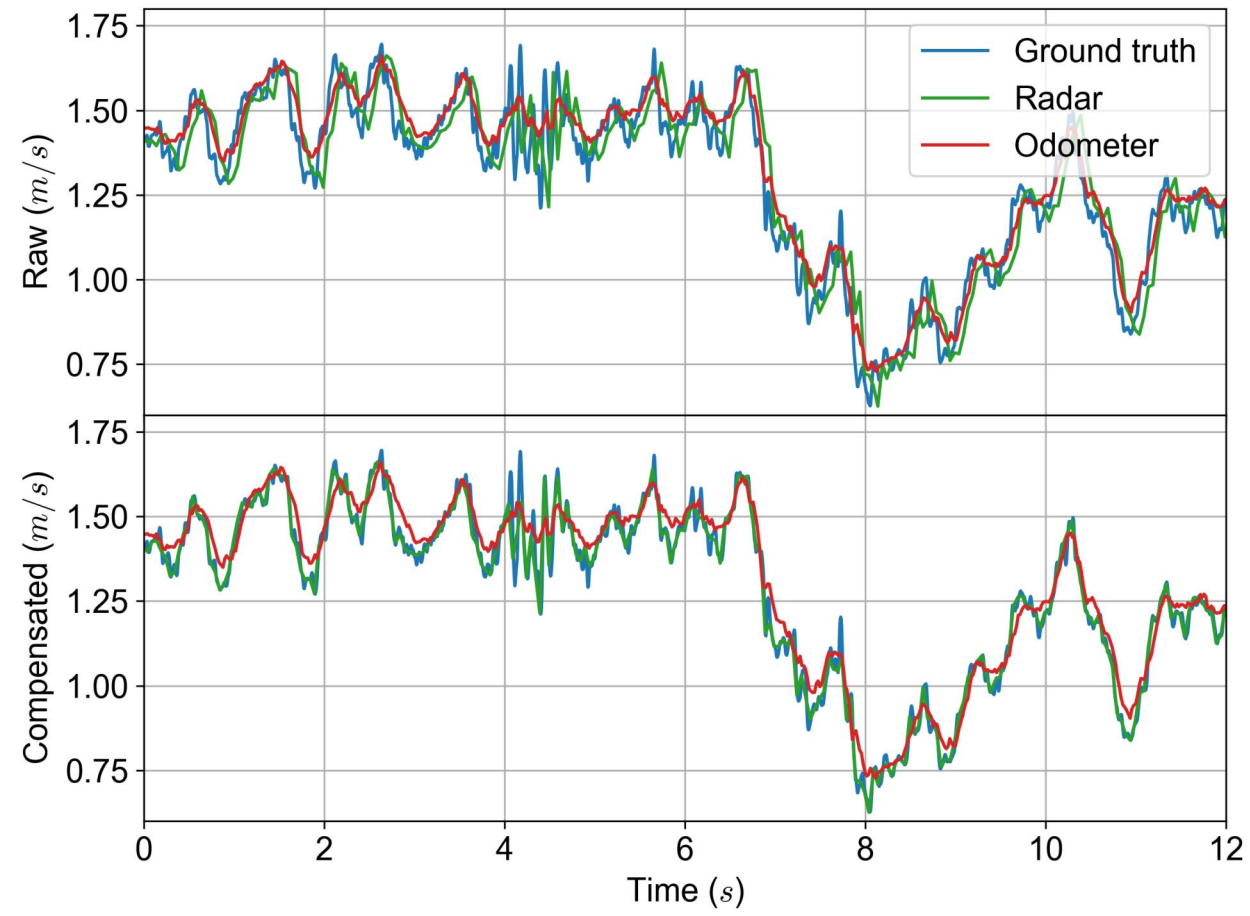

Fig. 7. Comparison of the speeds before (Raw) and after (Compensated) compensating the fixed time offsets for radar and ODO. The ODO and radar are originally delayed, as shown in the top figure.

## III. Ground-Truth System

A navigation-grade IMU is employed as the ground-truth system, which can provide high-rate (200 Hz) positions with centimeter-level accuracy, attitudes, and velocities. For sequences with good GNSS observations, we employ a tightly-coupled GNSS/INS integration to generate the ground truth, otherwise use a loosely-coupled map-matching (MM)/INS integration. Fig. 8 depicts the pipeline for generating the ground truth.

### A. GNSS/INS Integration

The GNSS raw observations from the OEM719 receiver, including pseudorange, carrier phase, and Doppler, are collected for post-processing GNSS-RTK/INS integrated navigation. The commercial software Inertial Explorer (IE) from NovAtel is adopted to process the GNSS data from the OEM719 and IMU data from the navigation-grade IMU. Besides, the IMU-rate (200 Hz) ground-truth positions, attitudes, and velocities are derived from the combined solution in IE, and thus they are all smoothed with centimeter-level accuracy for positions. Here, the centimeter-level accuracy is determined by the GNSS RTK positioning [7]. The attitude accuracy of the ground truth is about 0.01° for roll and pitch angles, and about 0.02° for yaw angle, according to the manufacturer's specification.

### B. MM/INS Integration

In GNSS-challenging urban canyons or GNSS-denied indoor parking lots, we cannot obtain centimeter-level RTK results, and the GNSS/INS integration may be unreliable in such scenarios. Hence, we employ an MM-based method to derive accurate ground truth in GNSS-denied environments.

We first build large-scale point-cloud maps in the building and parking scenarios, using a commercial LiDAR SLAM system, *i.e.*, the RS100i-MT from GoSLAM. The raw IMU and point clouds are collected using the RS100i-MT with delicately designed trajectories to ensure high-quality mapping. We also incorporate several good-quality RTK control points, which are well distributed, to maintain a higher absolute accuracy. The point-cloud maps are offline built by the commercial mapping software from GoSLAM, using the collected raw data and RTK control points. Hence, we can obtain high-quality point-cloud maps with a relative accuracy of about 1 cm, according to the specifications from GoSLAM. Besides, the built maps are aligned to the geodetic coordinate using the RTK control points with an absolute accuracy of about several centimeters. Note that the map points are expressed in a local geodetic coordinate, *i.e.* a local navigation frame [7], and no additional bias will be introduced. Fig. 9 exhibits the prebuilt point-cloud map of the second floor in the parking lot. The traffic lines are clear to distinguish, demonstrating the high quality of the map. We further examine the built map by checking the point coordinates at the control points, and the point residuals are about several centimeters, which indicates higher accuracy of the maps.

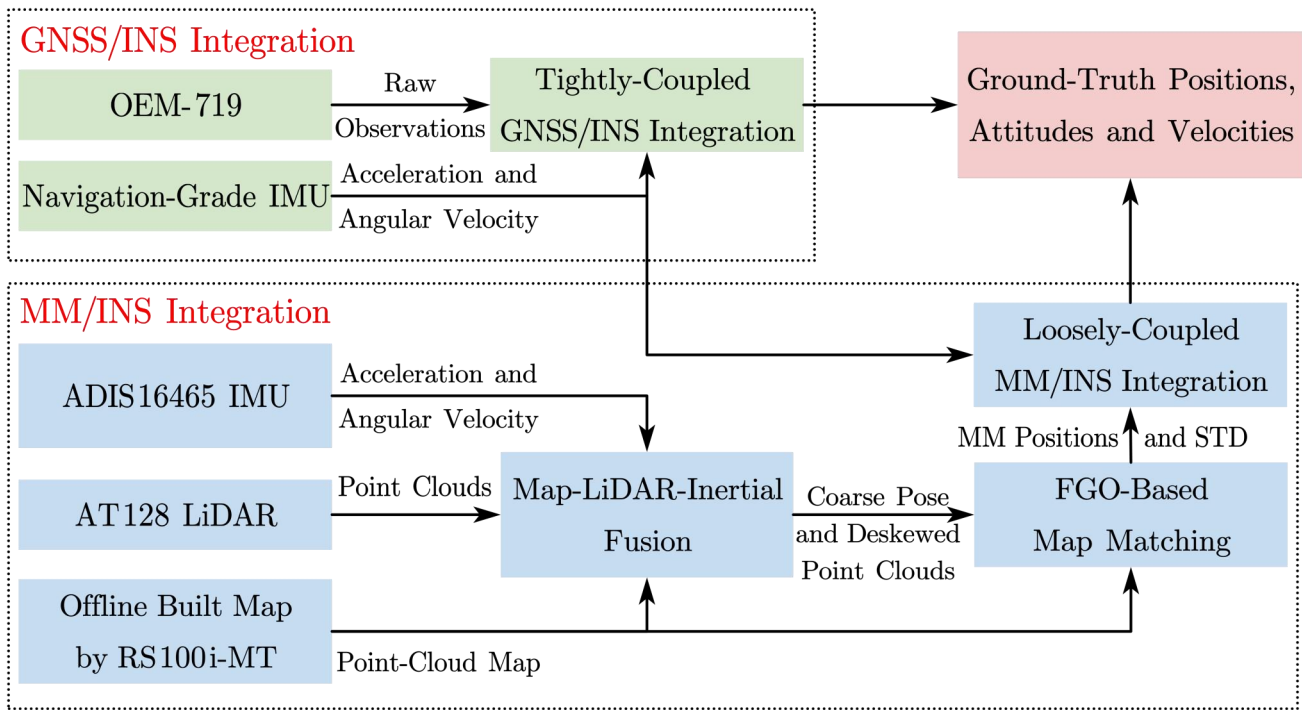

Fig. 8. An illustration of the employed ground-truth pipeline.

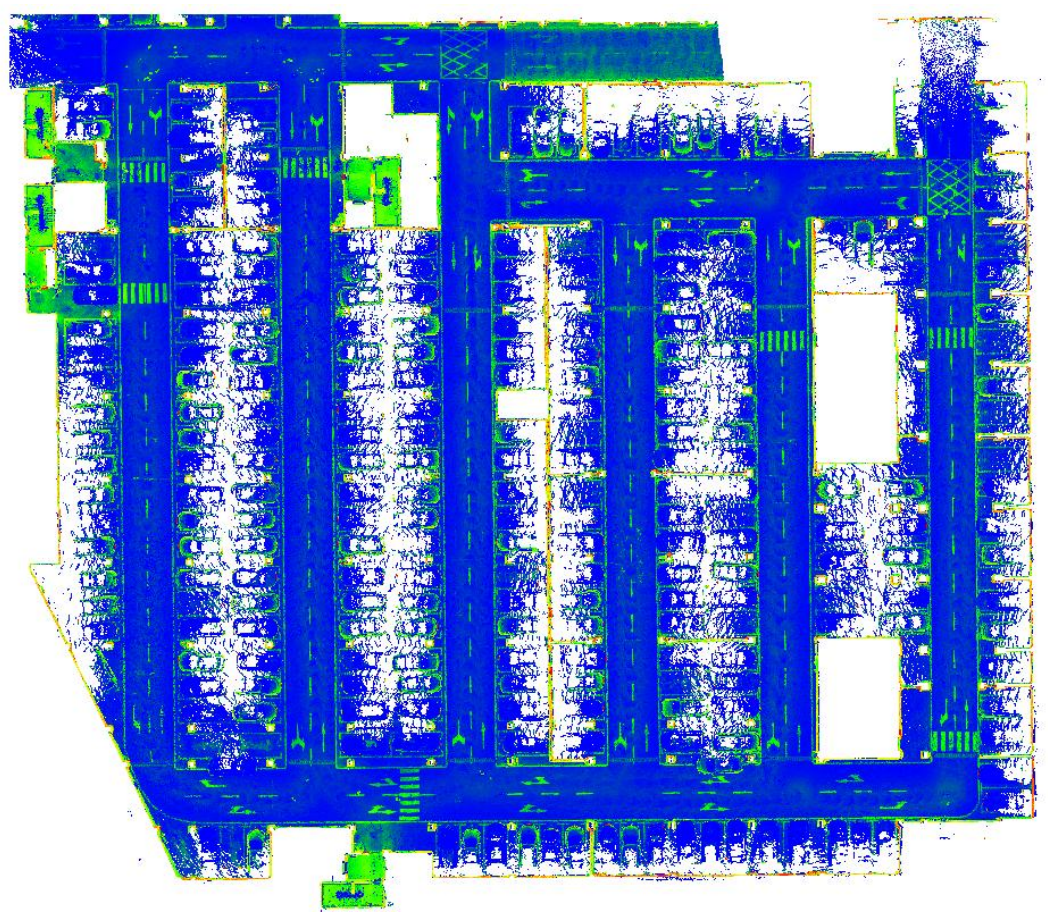

Fig. 9. The prebuilt point-cloud map of the second floor in the parking lot (used by three parking sequences). The traffic lines are clear to distinguish, which demonstrates the high quality of the map. This map has been cut for better visualization.

Next, we use a map-LiDAR-inertial fusion method with MM-based constraints from the AT128 LiDAR to obtain coarse position results and distortion-removed point clouds, as shown in Fig. 8. The employed map-LiDAR-inertial method is similar to BA-LINS [12], but further integrates MM-based constraints in a tightly-coupled form. Specifically, the MM-based constraint is derived from the point iterative closest point (ICP), which is very similar to FAST-LIO2 [37]. Note that the initial pose is manually set, and thus the loop-closure method is not involved.

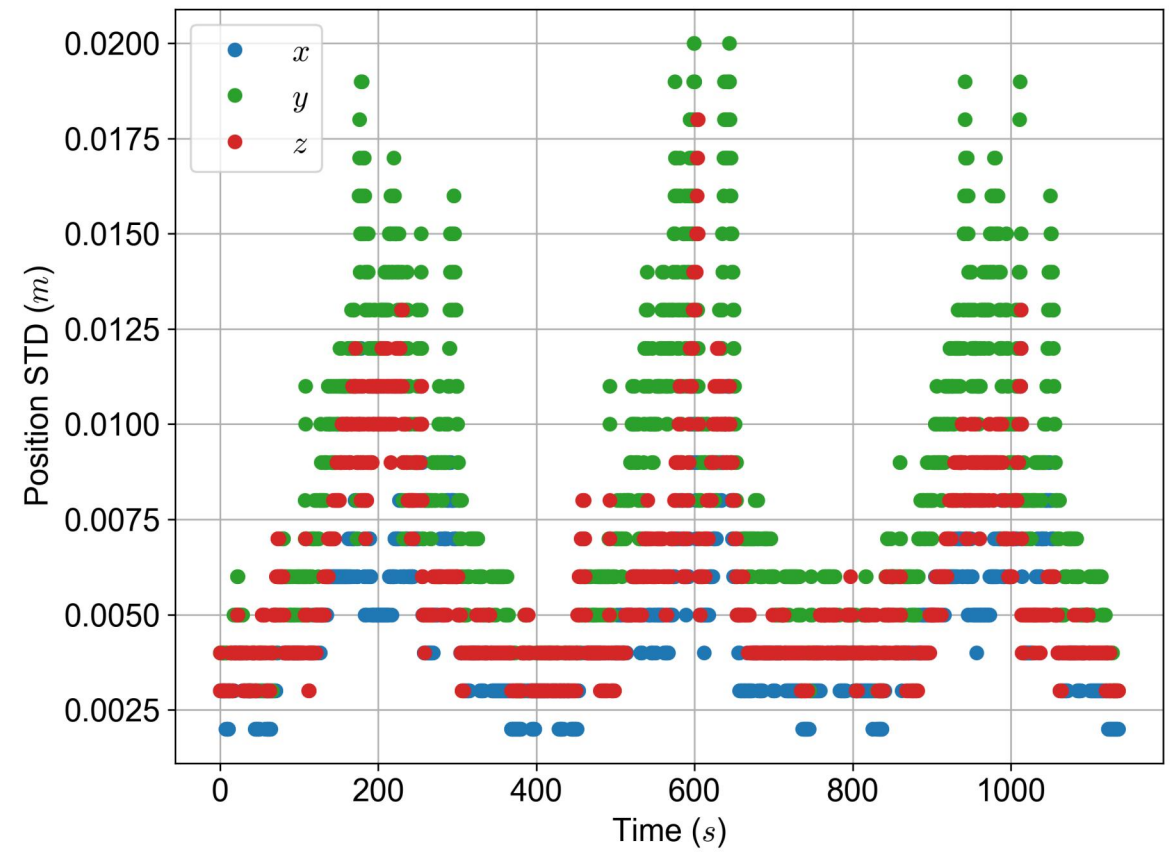

Fig. 10. The standard deviation (STD) of positions derived from the FGO-based MM in the sequence *parking00*. The resolution for the STD is 0.001 m. The STD is much larger in the indoor parking lot (at about 200, 600, and 1000 s), due to the smaller horizontal FOV of the AT128 LiDAR (120°).

Then, a similar factor-graph optimization (FGO)-based method is employed to derive epoch-independent MM results, without introducing epoch-coupling errors. The coarse pose, deskewed point clouds, and the prebuilt point-cloud map are used in the FGO-based map matching. Specifically, the Ceres Solver [38] is adopted to optimize the pose and calculate the pose covariance. Due to the higher absolute accuracy of the point-cloud maps, the uncertainty of the maps is not involved, similar to FAST-LIO2. Fig. 10 shows an example of the standard deviation (STD) of MM positions derived from the FGO-based MM , and it illustrates that the derived positions have a higher accuracy of about centimeter level.

Finally, the MM positions together with the covariance or the standard deviation (STD) and the navigation-grade IMU are integrated by a post-processing integration navigation software. As a result, we can obtain centimeter-level ground-truth positions even in GNSS-denied scenarios. Similarly, attitude accuracy is mainly determined by the navigation-grade IMU, with a yaw accuracy of about 0.02°.

## IV. Dataset Description

### A. Sequence Characteristics

The i2Nav-Robot dataset was collected in diverse scenarios in the campus of Wuhan University, which are the typical application scenarios for a commercial logistics and cleaning UGVs. Specifically, we collected ten sequences, including building, playground, parking, and street sequences. Fig. 11 shows the trajectories and scenarios of these ten sequences.

Table III shows the details about the ten sequences, and the total trajectory length is about 14204 seconds and 17060 meters. Both indoor, outdoor, and indoor-outdoor sequences are included in the proposed dataset. Specifically, the three building sequences contain outdoor environments with lots of tall buildings, and the *building01* sequence also contains an indoor parking lot. The three parking sequences are collected in an indoor multi-layer parking lot, while the indoor-outdoor *parking00* also contains an outdoor park scenario. The MM/INS integration is employed to derive the ground truth in building and parking scenarios due to the unavailable GNSS RTK. The *playground00* is collected in an open-sky playground with good GNSS-RTK results, and it can be served as the most basic benchmark. The three street sequences are the typical application scenarios for UGVs, with various buildings, trees, vehicles, and pedestrians.

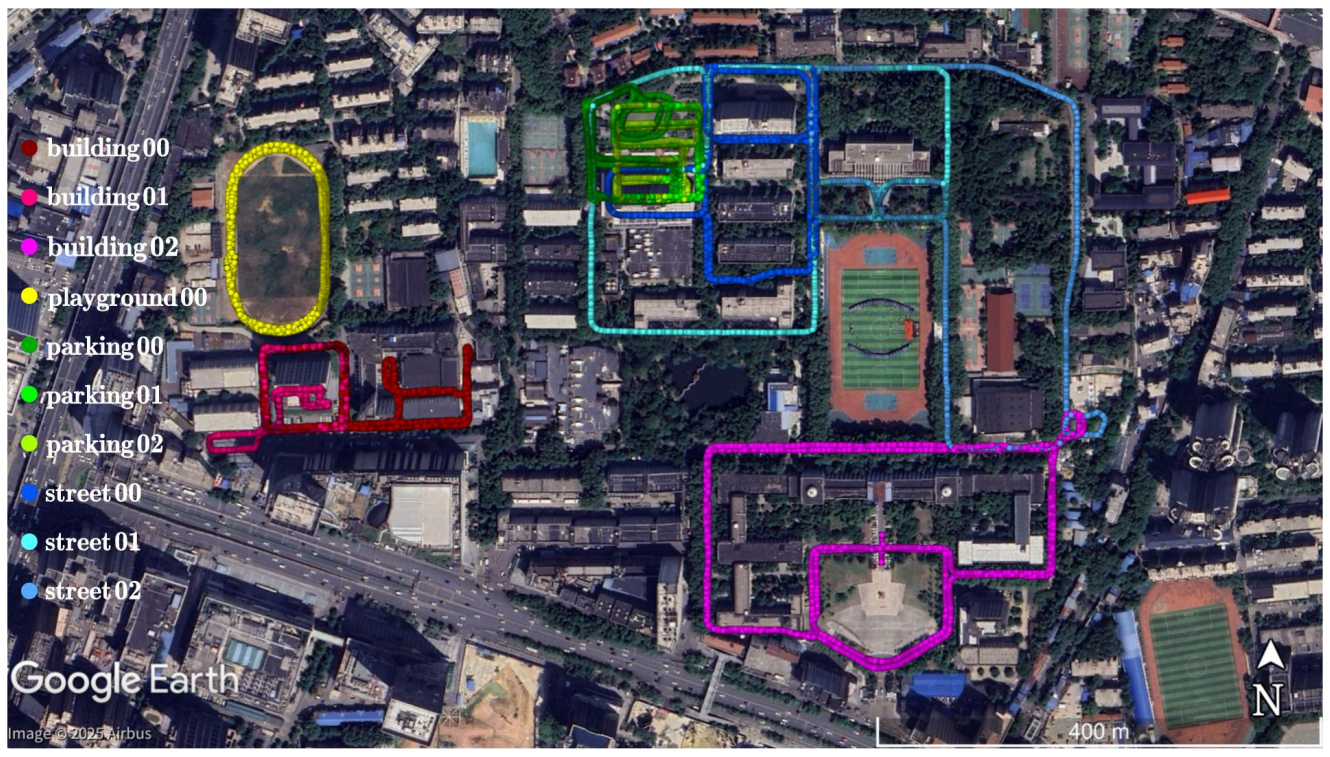

Fig. 11. Sequences overview in the i2Nav-Robot dataset. Each color denotes a different sequence.

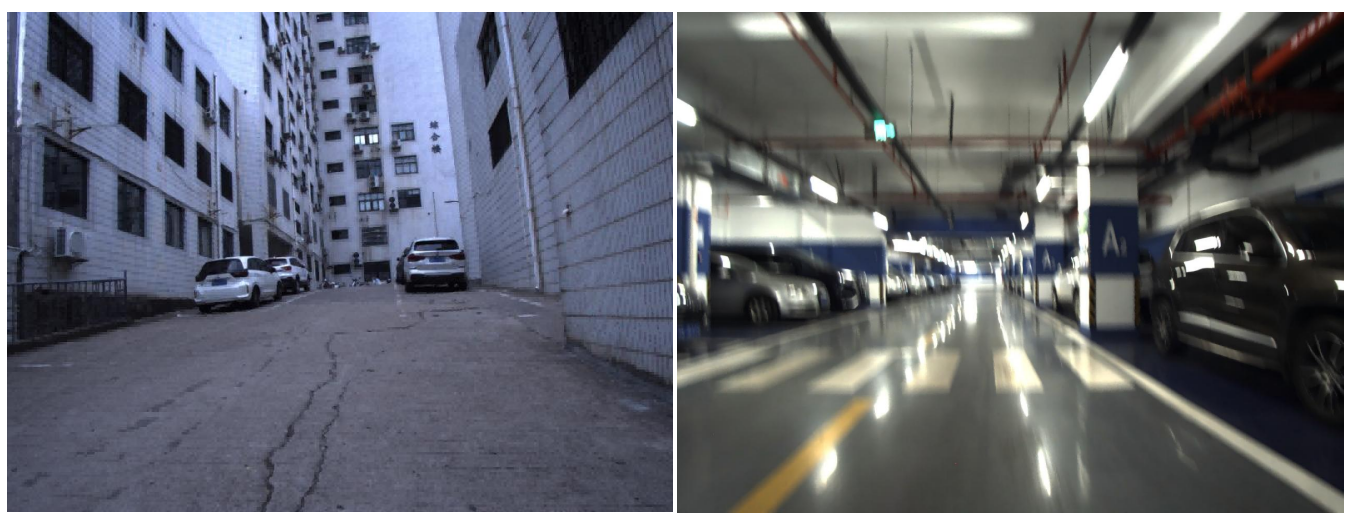

(a) GNSS-degenerated urban canyon. (b) Motion blur.

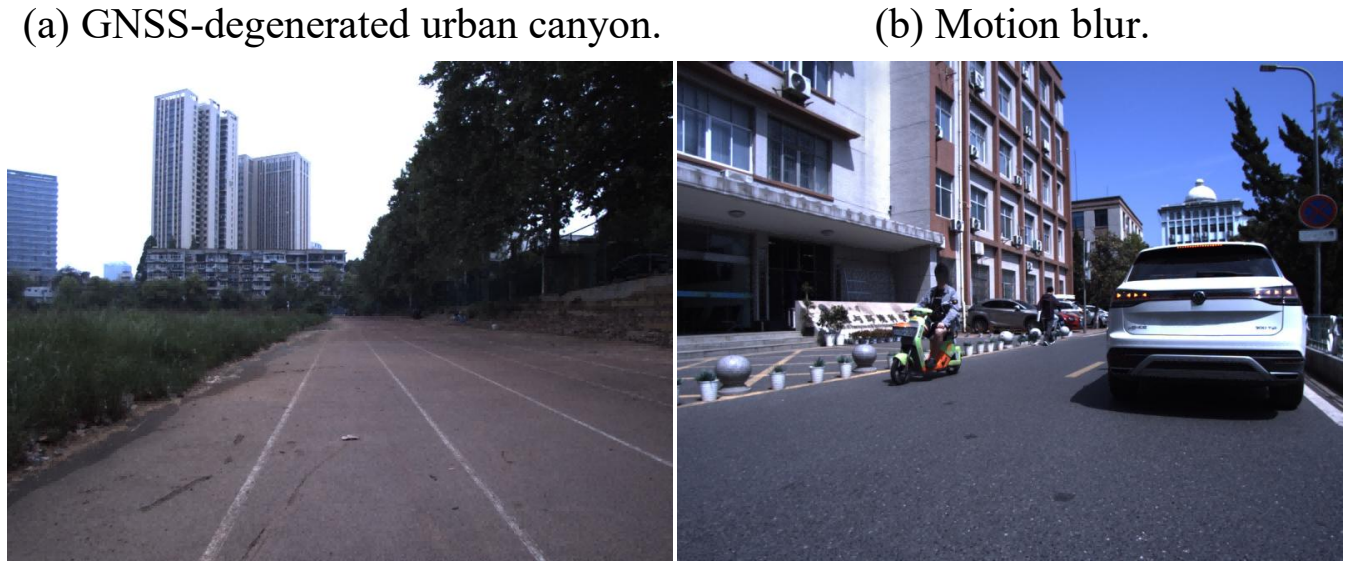

(c) Structureless scene. (d) Dynamic objects.

Fig. 12. Challenges for different sensors. (a) shows the GNSS-challenging environments. (b) shows the motion blur of camera. (c) shows a structureless scene for LiDAR and radar. (d) is a dynamic environment, which is challenging for perception sensors.

The diverse scenarios provide different challenges for multi-sensor fusion navigation methods. Fig. 12 depicts several challenging scenes for different sensors, such as GNSS-degenerated urban canyons in the three building sequences and the motion blur for cameras in the indoor parking sequences. In addition, the LiDAR and radar may be degraded in structureless scenes of the *playground00*, and dynamic objects may impact the performance of all perception sensors. These challenges are very meaningful to evaluate the robustness and reliability of a multi-sensor fusion method.

TABLE III
THE SEQUENCE DETAILS

| Sequence | Duration (s) | Trajectory (m) | Scene[1] | Challenges | GT |
|---|---|---|---|---|---|
| *building00* | 1201 | 1673 | OUT | GNSS degenerated | MM/INS |
| *building01* | 1369 | 1730 | IO | Illumination | MM/INS |
| *building02* | 2110 | 2522 | OUT | Dynamic | GNSS/INS |
| *parking00* | 1141 | 1583 | IO | Illumination | MM/INS |
| *parking01* | 1060 | 1224 | IN | Motion blur | MM/INS |
| *parking02* | 1257 | 1442 | IN | Motion blur | MM/INS |
| *playground00* | 1786 | 1711 | OUT | Structureless | GNSS/INS |
| *street00* | 1380 | 1668 | OUT | Dynamic | GNSS/INS |
| *street01* | 1520 | 1660 | OUT | Dynamic | GNSS/INS |
| *street02* | 1380 | 1847 | OUT | Dynamic | GNSS/INS |
| Total | 14204 | 17060 | - | - | - |

[1]IN denotes indoor; OUT denotes outdoor; IO denotes indoor and outdoor.

TABLE IV
THE TOPIC DETAILS IN THE ROS BAG FILE

| Topic | Rate (Hz) | Message Type |
|---|---|---|
| /adi/adis16465/imu | 200 | sensor_msgs::Imu |
| /insprobe/adis16465/imu | 200 | insprobe_msgs::Imu |
| /livox/mid360/imu | 200 | sensor_msgs::Imu |
| /ublox/f9p/fix | 1 | sensor_msgs::NavSatFix |
| /novatel/oem7/fix | 1 | sensor_msgs::NavSatFix |
| /avt_camera/left/image/compressed | 10 | sensor_msgs::CompressedImage |
| /avt_camera/right/image/compressed | 10 | sensor_msgs::CompressedImage |
| /livox/mid360/points | 10 | livox_ros_driver::CustomMsg |
| /hesai/at128/points | 10 | sensor_msgs::PointCloud2 |
| /continental/ars548/detection | 20 | sensor_msgs::PointCloud2 |
| /continental/ars548/object | 20 | sensor_msgs::PointCloud2 |
| /insprobe/ranger/odometer | 50 | insprobe_msgs::Odometer |

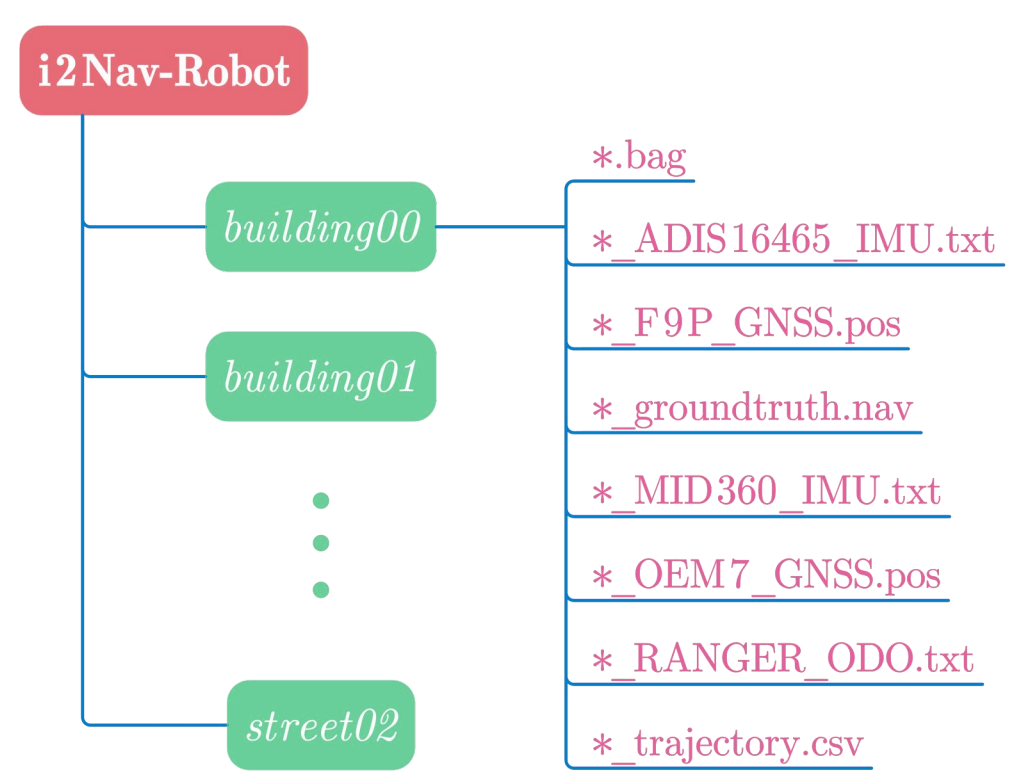

Fig. 13. The dataset organization, where * denotes the sequence name. There are six or eight files in the directory of each sequence, and GNSS files are not included in indoor sequences.

## B. Data Format

We provide all-in-one ROS bag files for all sequences, and thus the users can examine the dataset at once. Table IV shows the topic information, including topic name, frame rate, and message type. We also provide raw text files for IMU, GNSS, and ODO data, which can be utilized for GNSS/INS integration

TABLE V
THE ABSOLUTE ERRORS FOR ODOMETRY METHODS

| ARE (°) / ATE (m) | VINS-Mono | OpenVINS (Stereo) | DM-VIO | DLIO | FF-LINS | FAST-LIO2 | HKUST-RIO | FAST-LIVO2 | LE-VINS | R3LIVE |
|---|---|---|---|---|---|---|---|---|---|---|
| Sensors | V+I | V+I | V+I | L+I | L+I | L+I | R+I | L+V+I | L+V+I | L+V+I |
| *building00* | 0.48 / 3.95 | 1.34 / 3.36 | 0.52 / 2.45 | 0.90 / **0.39** | 2.19 / 2.32 | 1.20 / 0.68 | 13.04 / 14.74 | 0.91 / 1.10 | **0.39** / 1.17 | Failed |
| *building01* | 0.74 / 2.74 | 0.86 / 1.13 | 0.76 / 2.42 | 0.86 / 0.43 | 1.35 / 1.16 | 0.93 / 0.38 | 5.90 / 12.01 | 0.43 / **0.19** | **0.36** / 1.01 | Failed |
| *building02* | 1.21 / 6.51 | 0.34 / 1.48 | **0.34** / 10.31 | 8.56 / 9.62 | 2.82 / 6.73 | 0.76 / **1.46** | 32.14 / 65.69 | 1.04 / 1.65 | 0.79 / 2.35 | 1.70 / 4.49 |
| *parking00* | 1.14 / 7.53 | **0.33** / 0.88 | 0.37 / 1.19 | 0.73 / 0.46 | 2.48 / 1.94 | 0.45 / **0.25** | 8.58 / 15.67 | 0.56 / 0.26 | 0.50 / 0.98 | 1.12 / 0.59 |
| *parking01* | 3.67 / 3.86 | 0.34 / 0.81 | 0.83 / 2.33 | 0.58 / 0.15 | 1.69 / 1.06 | 1.21 / 0.44 | 7.79 / 9.51 | 0.65 / **0.17** | **0.30** / 0.69 | 0.91 / 0.39 |
| *parking02* | 3.13 / 3.24 | 2.34 / 3.47 | 0.85 / 1.31 | 1.08 / 0.56 | 1.64 / 1.31 | 2.07 / 1.46 | 6.26 / 10.40 | **0.65** / **0.18** | 0.69 / 0.79 | 1.37 / 0.66 |
| *playground00* | 0.51 / 2.59 | 0.92 / 1.62 | 0.85 / 8.07 | 8.21 / 0.44 | 1.43 / 1.71 | **0.41** / **0.31** | 7.84 / 13.88 | 0.49 / 0.53 | 0.84 / 1.34 | 0.52 / 0.62 |
| *street00* | 0.57 / 3.81 | 0.84 / 1.60 | Failed | 0.73 / **0.79** | 2.22 / 3.32 | 1.38 / 1.86 | 10.71 / 31.53 | 0.81 / 1.12 | **0.53** / 1.18 | 1.34 / 1.61 |
| *street01* | 1.07 / 8.71 | 0.37 / **1.11** | 0.64 / 3.44 | 8.84 / 4.31 | 1.75 / 4.20 | 0.72 / 1.28 | 7.45 / 14.83 | 1.15 / 2.45 | **0.34** / 1.65 | 1.43 / 2.36 |
| *street02* | 0.83 / 6.72 | **0.46** / 1.61 | 1.35 / 7.43 | 1.39 / 3.15 | 1.95 / 3.78 | 0.91 / 1.49 | 6.25 / 13.68 | 1.19 / 2.92 | 0.51 / 2.16 | 1.32 / **1.40** |
| RMS | 1.71 / 5.38 | 1.01 / 1.93 | Invalid | 4.74 / 3.50 | 2.00 / 3.23 | 1.11 / **1.12** | 12.97 / 25.91 | 0.83 / 1.42 | **0.55** / 1.43 | Invalid |

The bold result denotes the best among these systems. V: Vision; I: Inertial; L: LiDAR; R: Radar.

applications. Meanwhile, two types of ground-truth files are included in the dataset. The "*_groundtruth.nav" file includes the positions, attitudes, and velocities, and the "*_trajectory.csv" file contains the positions and attitude quaternions. The dataset organization is shown in Fig. 13, and the GNSS position files are not contained in the indoor sequences *parking01* and *parking02*. More details about the data format can be found in our online documents.

## V. EXPERIMENTS

We conduct exhaustive experiments to evaluate the quality of the presented i2Nav-Robot dataset. A total of 15 multi-sensor navigation methods are included for a comprehensive evaluation, and their configuration files are all open-sourced[1]. Meanwhile, we also carry on experiments to verify the accuracy of the spatial-temporal synchronization. The stereo camera, ADIS16465 IMU, AT128 LiDAR, ARS548 radar, and GNSS-RTK results from OEM719 are employed for the following experiments.

### *A. Evaluation of Odometry Methods*

Odometry methods, including VIO, LIO, radar-inertial odometry (RIO), and LiDAR-visual-inertial odometry (LVIO), determine the navigation accuracy in GNSS-denied environments. Hence, the performance of an odometry is crucial for autonomous UGVs. Three SOTA VIOs, including VINS-Mono [39], OpenVINS [40], and DM-VIO [41], and three LIOs, including DLIO [42], FF-LINS [33], and FAST-LIO2 [37], are adopted for evaluation. We have also tried SOTA RIOs, including EKF-RIO [43] and HKUST-RIO [44] on the presented dataset. However, EKF-RIO exhibits poor results with absolute translation error (ATE) larger than 50 m, mainly because it only uses Doppler measurements. Hence, the results of EKF-RIO are not presented. Meanwhile, three LVIOs using both camera and LiDAR, including FAST-LIVO2 [45], LE-VINS [34], and R3LIVE [46], are also included. The absolute rotation error (ARE) and ATE are calculated using evo [47]. The results of ten employed odometry methods are exhibited in Table V.

According to the results in Table V, only DM-VIO and R3LIVE fail to run on certain sequences, demonstrating the excellent quality of the dataset. Besides, VIO OpenVINS, LIO FAST-LIO2, and LVIO FAST-LIVO2, can achieve the best results on different sequences, which means that these methods are all correctly configured. Overall, FAST-LIO2 shows the best results among these methods in terms of the root-mean-square (RMS) errors. As the point clouds from the ARS548 radar are very sparse and noisy, HKUST-RIO exhibits the worst result in all sequences.

Specifically, the VIOs exhibit worse results than the LIOs, especially in street sequences with complex scenarios and in indoor parking sequences with frequently occurring motion blur. LiDAR-based methods yield smaller ATEs on the parking sequences, due to structured indoor environments, especially for methods using the frame-to-map association, including DLIO, FAST-LIO2, FAST-LIVO2, and R3LIVE. In contrast, these LiDAR-based methods exhibit much worse results on the streets sequence. The reason is that they cannot associate with their built map when large drifts occur in large-scale scenarios. Besides, nearly all these methods show worse results on the *building02*, due to the longest trajectory. We also notice that OpenVINS yields comparable or superior accuracy to LIOs or LVIOs on the *building02* and the three street sequences. The results demonstrate that current odometry methods can be further improved by fully integrating measurements from visual, LiDAR, radar, and inertial sensors.

### *B. Evaluation of GNSS-based Methods*

GNSS is the key factor to achieve accurate navigation in large-scale environments. Hence, we also evaluate GNSS-based multi-sensor fusion methods on the i2Nav-Robot dataset, including KF-GINS [48], OB-GINS [3], VINS-Fusion [49], IC-GVINS [50], and BA-GLINS (BA-LINS with GNSS) [12]. KF-GINS and OB-GINS are both GNSS/INS integrated navigation methods. VINS-Fusion and IC-GVINS are GNSS-visual-inertial navigation methods. BA-GLINS is a GNSS-LiDAR-inertial navigation method [12]. Note that we failed to run some SOTA multi-sensor fusion methods due to

[1]https://github.com/i2Nav-WHU/i2Nav-Robot-Baseline

TABLE VI
THE ABSOLUTE ERRORS FOR GNSS-BASED METHODS

| ARE (°) / ATE (m) | KF-GINS | OB-GINS | VINS-Fusion (Mono) | IC-GVINS | BA-GLINS |
|---|---|---|---|---|---|
| Sensors | G+I | G+I | G+V+I | G+V+I | G+L+I |
| *building00* | 66.08 / 13.91 | 1.66 / 8.46 | 36.60 / 14.04 | **0.30** / 0.72 | 0.53 / **0.50** |
| *building01* | 9.45 / 11.26 | 10.22 / 12.16 | 9.12 / 10.83 | 0.36 / 0.33 | **0.26 / 0.17** |
| *building02* | 1.75 / 2.72 | 1.37 / 2.80 | 8.27 / 2.80 | 0.23 / 0.23 | **0.22 / 0.20** |
| *parking00* | 34.16 / 23.28 | 24.75 / 23.87 | 24.20 / 23.25 | **0.49** / 0.74 | 0.54 / **0.27** |
| *playground00* | 0.56 / 0.40 | 1.17 / 0.40 | 9.40 / 0.31 | **0.15 / 0.11** | 0.23 / 0.12 |
| *street00* | 1.28 / 0.40 | 1.78 / 0.40 | 1.42 / 0.29 | **0.15 / 0.10** | 0.17 / 0.11 |
| *street01* | 1.20 / 0.48 | 1.06 / 0.48 | 0.89 / 0.29 | 0.46 / 0.29 | **0.24 / 0.10** |
| *street02* | 0.70 / 0.42 | 1.51 / 0.42 | 4.11 / 0.26 | **0.19** / 0.12 | 0.33 / **0.12** |
| RMS | 26.53 / 10.43 | 9.55 / 9.98 | 16.53 / 10.39 | **0.32** / 0.41 | 0.34 / **0.24** |

G: GNSS; V: Vision; I: Inertial; L: LiDAR.

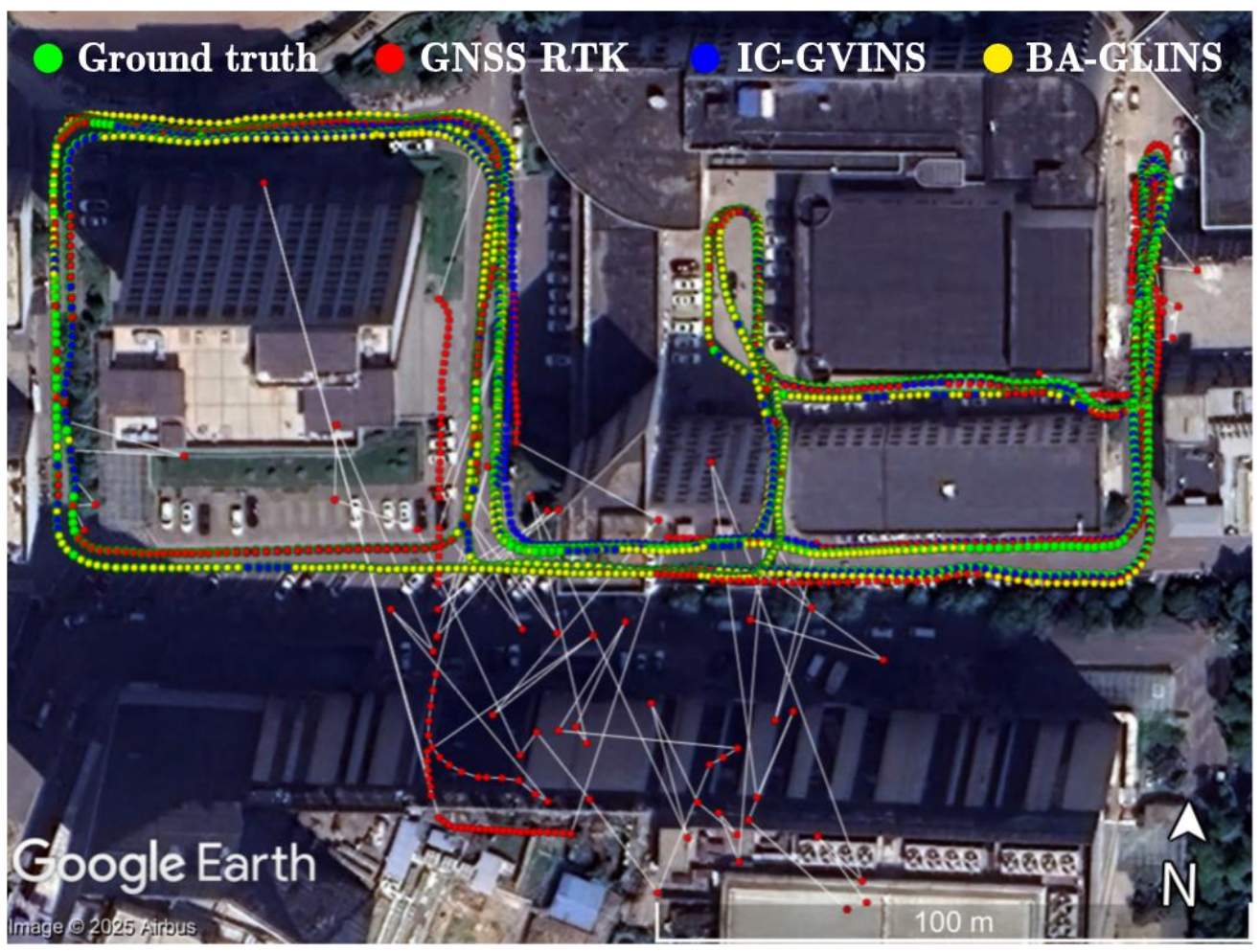

Fig. 14. Comparison of the results on the GNSS-degenerated sequence *builing00*. Due to worse results, KF-GINS, OB-GINS, and VINS-Fusion are not included in this figure for better visualization.

their complex configurations, such as MINS [51] and GICI-LIB [52], and thus their results are not included.

Table VI shows the absolute errors of these GNSS-based methods. The *building00* is a GNSS-degenerated sequence, and the GNSS RTK exhibits wrongly fixed positions, as shown in Fig. 14. Besides, the *building01* and *parking00* are all indoor-outdoor sequences, in which the RTK is interrupted for a long time. As a result, all systems show worse results on the three sequences. Nevertheless, BA-GLINS almost achieves a decimeter-level positioning accuracy, while IC-GVINS achieves a sub-meter accuracy. This is because BA-GLINS and IC-GVINS employ an effective outlier-culling method for RTK, while other systems use all RTK results despite outliers. In particular, BA-GLINS yield the smallest RMS result, as the LiDAR is more precise than the vision sensor. The results demonstrate that KF-GINS, OB-GINS, and VINS-Fusion are more suitable for applications with good-quality GNSS positioning, while IC-GVINS and BA-GLINS are applicable in GNSS-challenging scenarios.

The results in Table VI also demonstrate the higher accuracy of the ground-truth pose, as the ATEs of BA-GLINS and IC-GVINS in some sequences are about 0.1 m. Note that the real-time RTK results from the OEM719 are employed for experiments. In contrast, the post-processing tightly-coupled GNSS/INS integration is adopted to derive ground-truth positions, which are far more accurate than the RTK results from OEM719. Hence, an ATE of 0.1 m can reflect the higher accuracy of the ground-truth pose.

TABLE VII
ESTIMATED CAMERA-IMU EXTRINSIC PARAMETERS BY LE-VINS

| Sequence | Translation (m) | | | Rotation (°) | | |
|---|---|---|---|---|---|---|
| | x | y | z | x | y | z |
| *building00* | 0.088 | -0.195 | -0.040 | 89.974 | -0.184 | 89.663 |
| *building01* | 0.087 | -0.195 | -0.043 | 89.972 | -0.177 | 89.700 |
| *building02* | 0.091 | -0.195 | -0.034 | 89.946 | -0.171 | 89.694 |
| *parking00* | 0.091 | -0.191 | -0.036 | 89.936 | -0.176 | 89.707 |
| *parking01* | 0.080 | -0.178 | -0.067 | 89.887 | -0.171 | 89.679 |
| *parking02* | 0.082 | -0.201 | -0.059 | 89.870 | -0.181 | 89.782 |
| *playground00* | 0.086 | -0.199 | ***-0.018*** | 89.974 | -0.187 | 89.667 |
| *street00* | 0.088 | -0.202 | -0.041 | 89.955 | -0.172 | 89.677 |
| *street01* | 0.090 | -0.194 | -0.036 | 89.956 | -0.175 | 89.711 |
| *street02* | 0.090 | -0.202 | -0.042 | 89.960 | -0.168 | 89.684 |
| Mean[1] | 0.087 ±0.004 | -0.195 ±0.007 | -0.042 ±0.013 | 89.943 ±0.034 | -0.176 ±0.006 | 89.696 ±0.032 |

[1]The mean value and the standard deviation. The estimated z-axis translation is much larger on the sequence *playground00* due to the weak observability.

## C. *Evaluation of the Synchronization*

Spatial-temporal synchronization is crucial for a multi-sensor fusion navigation system, especially for frequently used LiDAR, camera, and IMU sensors. The IMU can act as a converter of spatial-temporal parameters, and thus we evaluate the camera-IMU and LiDAR-IMU spatial-temporal synchronization to examine the dataset.

### 1) *Camera-IMU Synchronization*

The extrinsic parameters are not explicitly calibrated on the i2Nav-Robot dataset, mainly for rotation parameters. Hence, we provide the online estimated rotation parameters by LE-VINS (VIO mode) [34]. Table VII shows the results on each sequence, including the translation and rotation parts, and Fig. 15 shows the estimated parameters on the indoor-outdoor sequence *building01*.

The camera-IMU translation parameters obtained from the mechanical structure are (0.09, -0.2, -0.054) meters, which is close to the results in Table VII. In terms of the mean value (0.087, -0.195, -0.042) meters, the estimated translation parameters have an accuracy better than 1.5 cm. Meanwhile, both the translation and rotation parameters exhibit good consistency on different sequences, demonstrating the superior quality of the dataset. We notice the z-axis translation exhibits a large deviation on the *playground00*. This is because the ground is flat, lacking motion along the vertical axis, and thus this parameter is weakly observable. In short, the results in Table VII show that the online estimated translation and rotation parameters have a repeatability of better than 1.5 cm and 0.05°, respectively, which demonstrates that the online calibrations can also yield satisfactory accuracy.

We do not evaluate the time synchronization for camera and IMU on each sequence, due to the time-varying exposure time. Nevertheless, the estimated time-delay parameter by LE-VINS on the indoor-outdoor sequence *building01* is shown in Fig 15.

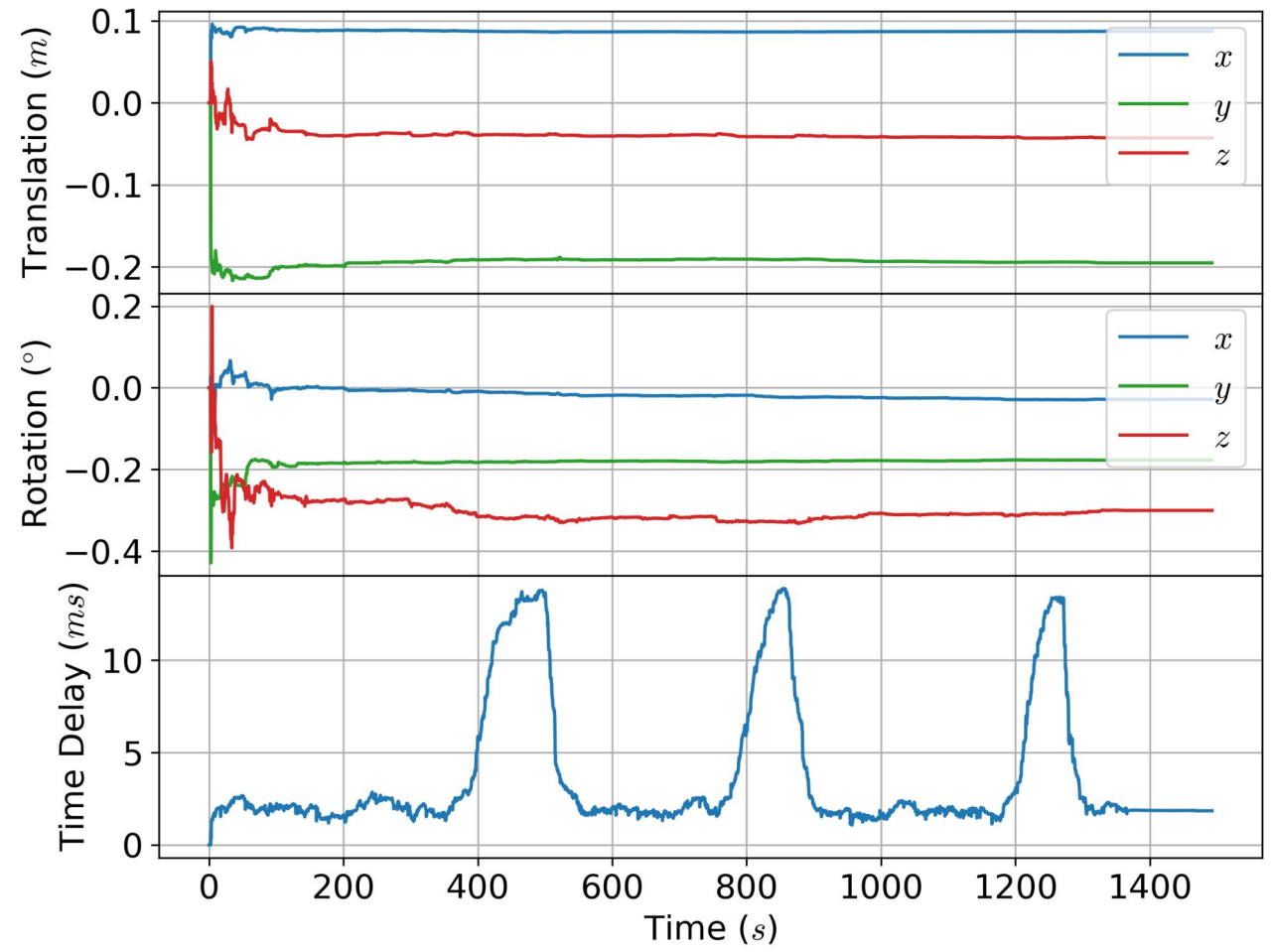

Fig. 15. Estimated camera-IMU extrinsic and time-delay parameters by LE-VINS on the indoor-outdoor sequence, *i.e.*, *building01*. Here, only the tiny rotation parameters are reserved for better visualization. The maximum exposure time for camera is set to 30 ms, and thus the estimated time-delay parameter is about 15 ms when in the indoor environments.

TABLE VIII
ESTIMATED LIDAR-IMU EXTRINSIC PARAMETERS BY FF-LINS

| Sequence | Translation (m) | | | Rotation (°) | | | Time delay (ms) |
|---|---|---|---|---|---|---|---|
| | x | y | z | x | y | z | |
| *building00* | 0.041 | -0.014 | -0.081 | 179.913 | 0.459 | 359.572 | 0.39 |
| *building01* | 0.038 | -0.023 | -0.099 | 179.918 | 0.407 | 359.543 | 0.20 |
| *building02* | 0.054 | -0.022 | -0.076 | 179.924 | 0.403 | 359.603 | 0.40 |
| *parking00* | 0.056 | -0.022 | -0.078 | 179.917 | 0.390 | 359.593 | 0.36 |
| *parking01* | 0.042 | -0.018 | -0.077 | 179.927 | 0.464 | 359.523 | 0.42 |
| *parking02* | 0.042 | -0.02 | -0.099 | 179.916 | 0.457 | 359.543 | ***0.84*** |
| *playground00* | 0.037 | -0.009 | ***-0.161*** | 179.931 | 0.462 | 359.542 | 0.49 |
| *street00* | 0.049 | -0.024 | -0.091 | 179.902 | 0.448 | 359.609 | 0.39 |
| *street01* | 0.050 | -0.021 | -0.059 | 179.935 | 0.451 | 359.631 | 0.38 |
| *street02* | 0.055 | -0.027 | -0.061 | 179.927 | 0.449 | 359.569 | 0.56 |
| Mean | 0.046 ±0.007 | -0.020 ±0.005 | -0.088 ±0.028 | 179.921 ±0.009 | 0.439 ±0.026 | 359.573 ±0.034 | 0.44 ±0.16 |

The abnormal estimated time-delay parameter on the sequence *parking02* is caused by the drifts of crystal oscillators in GNSS-denied indoor environments. Besides, the estimated z-axis translation is much larger on the sequence *playground00*, because of the insufficient observability.

Note that the camera-IMU time-delay parameter in LE-VINS is modeled as a random walk, as it is time varying. According to our configurations, the maximum exposure time for the camera is set to 30 ms, and thus the estimated time-delay parameter is about 15 ms (half of the exposure time) when in indoor environments. As exhibited in Fig. 15, we can easily determine that the robot walks into the indoor parking lot three times, and LE-VINS can accurately estimate the exposure time with an accuracy of about one millisecond. Here, the fixed time-delay parameter for camera and IMU in the outdoor environments is about 2.0 ms, due to the internal filters of ADIS16465. Nevertheless, the results demonstrate that the time-synchronization accuracy for camera and IMU is better than 1 ms.

*2) LiDAR-IMU Synchronization*

Similarly, we evaluate the LiDAR-IMU synchronization by analyzing the estimated extrinsic parameters by FF-LINS [33]. According to the mechanical structure, the LiDAR-IMU translation parameters are (0.042, -0.018, -0.063) meters. As shown in Table VIII, the estimated translation error for the x-axis and y-axis is within 1.0 cm, while 2.5 cm for the z-axis. This is caused by the smaller FOV (25°) of the AT128 LiDAR in the vertical direction, and thus the z-axis translation is weakly observable. Meanwhile, the rotation parameters have a repeatability of about 0.05°. In short, the results in Table VIII illustrate that the provided extrinsic parameters have a higher accuracy.

The LiDAR and IMU sensors are accurately synchronized using hardware-based methods. According to the results in Table VIII, the mean value of the estimated time-delay parameter between LiDAR and IMU is about 0.44 ms. This tiny time may be caused by the internal processes of LiDAR and IMU. Besides, we notice the time-delay parameter is much larger in the sequence *parking02*. The three parking sequences are collected one by one, and thus the crystal oscillators may also drift when in the indoor environments without the absolute time correction from the GNSS pulse-per-second (PPS) signal. Nevertheless, the LiDAR-IMU time synchronization has been proven to have accuracy better than 1 ms.

## VI. CONCLUSIONS

In this paper, we present a large-scale indoor-outdoor dataset, i2Nav-Robot, to satisfy the urgent demand for high-quality UGV datasets. The newest front-view LiDAR, 360-degree LiDAR, and 4D MMW radar are utilized in the dataset to follow the development of the modern mass-produced autonomous UGVs. Besides, multiple sensors are precisely synchronized or calibrated to achieve a time-synchronization accuracy better than one millisecond. Ten large-scale sequences with a total length of more than 17 kilometers are collected in both indoor, outdoor, and indoor-outdoor environments, with various challenging scenarios for different sensors. The high-rate and accurate ground-truth positions, attitudes, and velocities are provided for all sequences. Meanwhile, the proposed dataset is examined by 15 multi-sensor fusion methods and yields superior data quality.

However, the presented dataset currently lacks semantic annotations and ground-truth maps, which are crucial for intelligent transportation systems. Besides, all sequences were captured under clear weather conditions, with no representation of extreme weather such as rain or snow. The data collection was also conducted almost exclusively during spring, thereby missing the seasonal variations in vegetation that occur in summer, autumn, and winter. Hence, future efforts to extend the i2Nav-Robot dataset may incorporate additional data types, such as semantic annotations, and gather more diverse sequences across various weather conditions and seasons.

The i2Nav-Robot dataset is primarily designed for autonomous UGV applications, such as logistics, cleaning, and patrol, in closed environments like campuses and parks. It has demonstrated applicability in GNSS/INS integrated navigation, SLAM, and multi-sensor fusion navigation. Nevertheless, its potential in multi-sensor fusion navigation remains to be further explored. The following future research directions are proposed based on the dataset:

1) Constructing a large-scale RGB-rendered point-cloud map by integrating LiDAR, visual, and inertial sensors.

2) Developing tightly-coupled visual-radar-inertial fusion methods for reliable positioning in complex environments, without relying on the expensive LiDAR.

3) Enabling crowd-sourced mapping in large-scale environments using multi-sensor fusion, particularly leveraging the parking and street sequences.

4) Training deep-learning-based navigation methods with the accurate ground truth provided by the dataset.

## References

[1] X. Zhang *et al.*, "OpenMPD: An Open Multimodal Perception Dataset for Autonomous Driving," *IEEE Trans. Veh. Technol.*, vol. 71, no. 3, pp. 2437–2447, Mar. 2022, doi: 10.1109/TVT.2022.3143173.

[2] R. Siegwart, I. R. Nourbakhsh, and D. Scaramuzza, *Introduction to autonomous mobile robots*, 2nd ed. in Intelligent robotics and autonomous agents. Cambridge, Mass: MIT Press, 2011.

[3] H. Tang, T. Zhang, X. Niu, J. Fan, and J. Liu, "Impact of the Earth Rotation Compensation on MEMS-IMU Preintegration of Factor Graph Optimization," *IEEE Sens. J.*, vol. 22, no. 17, pp. 17194–17204, Sep. 2022, doi: 10.1109/JSEN.2022.3192552.

[4] T. Zhang, M. Yuan, L. Wang, H. Tang, and X. Niu, "A Robust and Efficient IMU Array/GNSS Data Fusion Algorithm," *IEEE Sens. J.*, vol. 24, no. 16, pp. 26278–26289, Aug. 2024, doi: 10.1109/JSEN.2024.3418383.

[5] P. J. G. Teunissen and O. Montenbruck, Eds., *Springer Handbook of Global Navigation Satellite Systems*. Cham: Springer International Publishing, 2017. doi: 10.1007/978-3-319-42928-1.

[6] J. Li, X. Xu, Z. Liu, S. Yuan, M. Cao, and L. Xie, "AEOS: Active Environment-aware Optimal Scanning Control for UAV LiDAR-Inertial Odometry in Complex Scenes," *ISPRS J. Photogramm. Remote Sens.*, vol. 232, pp. 476–491, Feb. 2026, doi: 10.1016/j.isprsjprs.2026.01.006.

[7] P. D. Groves, *Principles of GNSS, inertial, and multisensor integrated navigation systems*. Boston: Artech House, 2008.

[8] Z. Zhang, X. Niu, H. Tang, Q. Chen, and T. Zhang, "GNSS/INS/ODO/wheel angle integrated navigation algorithm for an all-wheel steering robot," *Meas. Sci. Technol.*, vol. 32, no. 11, p. 115122, Nov. 2021, doi: 10.1088/1361-6501/ac17fb.

[9] L. Wang, X. Niu, T. Zhang, H. Tang, and Q. Chen, "Accuracy and robustness of ODO/NHC measurement models for wheeled robot positioning," *Measurement*, vol. 201, p. 111720, Sep. 2022, doi: 10.1016/j.measurement.2022.111720.

[10] L. Wang, H. Tang, T. Zhang, Y. Wang, Q. Zhang, and X. Niu, "PO-KF: A Pose-Only Representation-based Kalman Filter for Visual Inertial Odometry," *IEEE Internet Things J.*, pp. 1–20, 2025, doi: 10.1109/JIOT.2025.3526811.

[11] J. Li, S. Yuan, M. Cao, T.-M. Nguyen, K. Cao, and L. Xie, "HCTO: Optimality-aware LiDAR inertial odometry with hybrid continuous time optimization for compact wearable mapping system," *ISPRS J. Photogramm. Remote Sens.*, vol. 211, pp. 228–243, May 2024, doi: 10.1016/j.isprsjprs.2024.04.004.

[12] H. Tang, T. Zhang, L. Wang, M. Yuan, and X. Niu, "BA-LINS: A Frame-to-Frame Bundle Adjustment for LiDAR-Inertial Navigation," *IEEE Trans. Intell. Transp. Syst.*, vol. 26, no. 5, pp. 6621–6634, May 2025, doi: 10.1109/TITS.2024.3524569.

[13] X. Shi *et al.*, "Are we ready for service robots? the openloris-scene datasets for lifelong slam," in *2020 IEEE international conference on robotics and automation (ICRA)*, IEEE, 2020, pp. 3139–3145.

[14] Y. Tian *et al.*, "Resilient and Distributed Multi-Robot Visual SLAM: Datasets, Experiments, and Lessons Learned," in *2023 IEEE/RSJ International Conference on Intelligent Robots and Systems (IROS)*, Oct. 2023, pp. 11027–11034. doi: 10.1109/IROS55552.2023.10342377.

[15] T.-M. Nguyen *et al.*, "MCD: Diverse large-scale multi-campus dataset for robot perception," in *Proceedings of the IEEE/CVF conference on computer vision and pattern recognition*, 2024, pp. 22304–22313. Accessed: Aug. 09, 2025. [Online]. Available: http://openaccess.thecvf.com/content/CVPR2024/html/Nguyen_MCD_Diverse_Large-Scale_Multi-Campus_Dataset_for_Robot_Perception_CVPR_2024_paper.html

[16] Y. Liu *et al.*, "BotanicGarden: A High-Quality Dataset for Robot Navigation in Unstructured Natural Environments," *IEEE Robot. Autom. Lett.*, vol. 9, no. 3, pp. 2798–2805, Mar. 2024, doi: 10.1109/LRA.2024.3359548.

[17] Y. Tao, M. Á. Muñoz-Bañón, L. Zhang, J. Wang, L. F. T. Fu, and M. Fallon, "The Oxford Spires Dataset: Benchmarking large-scale LiDAR-visual localisation, reconstruction and radiance field methods," *Int. J. Robot. Res.*, p. 02783649251369905, Sep. 2025, doi: 10.1177/02783649251369905.

[18] J. Yin, A. Li, T. Li, W. Yu, and D. Zou, "M2DGR: A Multi-Sensor and Multi-Scenario SLAM Dataset for Ground Robots," *IEEE Robot. Autom. Lett.*, vol. 7, no. 2, pp. 2266–2273, Apr. 2022, doi: 10.1109/LRA.2021.3138527.

[19] T. Pire, M. Mujica, J. Civera, and E. Kofman, "The Rosario dataset: Multisensor data for localization and mapping in agricultural environments," *Int. J. Robot. Res.*, vol. 38, no. 6, pp. 633–641, May 2019, doi: 10.1177/0278364919841437.

[20] D. Zhang *et al.*, "Towards Robust Sensor-Fusion Ground SLAM: A Comprehensive Benchmark and A Resilient Framework," in *2025 IEEE/RSJ International Conference on Intelligent Robots and Systems (IROS)*, Oct. 2025, pp. 8894–8901. doi: 10.1109/IROS60139.2025.11247507.

[21] Y. Zhuang, B. Wang, J. Huai, and M. Li, "4D iRIOM: 4D Imaging Radar Inertial Odometry and Mapping," *IEEE Robot. Autom. Lett.*, vol. 8, no. 6, pp. 3246–3253, Jun. 2023, doi: 10.1109/LRA.2023.3266669.

[22] Z. Yan, L. Sun, T. Krajník, and Y. Ruichek, "EU Long-term Dataset with Multiple Sensors for Autonomous Driving," in *2020 IEEE/RSJ International Conference on Intelligent Robots and Systems (IROS)*, 2020, pp. 10697–10704. doi: 10.1109/IROS45743.2020.9341406.

[23] J. Zhang *et al.*, "NTU4DRadLM: 4D Radar-Centric Multi-Modal Dataset for Localization and Mapping," in *2023 IEEE 26th International Conference on Intelligent Transportation Systems (ITSC)*, Sep. 2023, pp. 4291–4296. doi: 10.1109/ITSC57777.2023.10422606.

[24] N. Carlevaris-Bianco, A. K. Ushani, and R. M. Eustice, "University of Michigan North Campus long-term vision and lidar dataset," *Int. J. Robot. Res.*, vol. 35, no. 9, pp. 1023–1035, 2016.

[25] D. Zhang *et al.*, "Towards Robust Sensor-Fusion Ground SLAM: A Comprehensive Benchmark and A Resilient Framework," Jul. 11, 2025, *arXiv*: arXiv:2507.08364. doi: 10.48550/arXiv.2507.08364.

[26] J. Jiao *et al.*, "FusionPortable: A Multi-Sensor Campus-Scene Dataset for Evaluation of Localization and Mapping Accuracy on Diverse Platforms," in *2022 IEEE/RSJ International Conference on Intelligent Robots and Systems (IROS)*, Oct. 2022, pp. 3851–3856. doi: 10.1109/IROS47612.2022.9982119.

[27] H. Wei *et al.*, "FusionPortableV2: A unified multi-sensor dataset for generalized SLAM across diverse platforms and scalable environments," *Int. J. Robot. Res.*, p. 02783649241303525, Dec. 2024, doi: 10.1177/02783649241303525.

[28] H. Li *et al.*, "MARS-LVIG dataset: A multi-sensor aerial robots SLAM dataset for LiDAR-visual-inertial-GNSS fusion," *Int. J. Robot. Res.*, p. 02783649241227968, Jan. 2024, doi: 10.1177/02783649241227968.

[29] M. Burri *et al.*, "The EuRoC micro aerial vehicle datasets," *Int. J. Robot. Res.*, vol. 35, no. 10, pp. 1157–1163, Sep. 2016, doi: 10.1177/0278364915620033.

[30] *IEEE Standard for a Precision Clock Synchronization Protocol for Networked Measurement and Control Systems*, IEEE Std 1588-2008 (Revision of IEEE Std 1588-2002), Jul. 24, 2008. doi: 10.1109/IEEESTD.2008.4579760.

[31] A. Kaehler and G. Bradski, *Learning OpenCV 3: computer vision in C++ with the OpenCV library*. O'Reilly Media, Inc., 2016.

[32] E. Olson, "AprilTag: A robust and flexible visual fiducial system," in *2011 IEEE International Conference on Robotics and Automation*, May 2011, pp. 3400–3407. doi: 10.1109/ICRA.2011.5979561.

[33] H. Tang, T. Zhang, X. Niu, L. Wang, L. Wei, and J. Liu, "FF-LINS: A Consistent Frame-to-Frame Solid-State-LiDAR-Inertial State Estimator," *IEEE Robot. Autom. Lett.*, vol. 8, no. 12, pp. 8525–8532, Dec. 2023, doi: 10.1109/LRA.2023.3329625.

[34] H. Tang, X. Niu, T. Zhang, L. Wang, and J. Liu, "LE-VINS: A Robust Solid-State-LiDAR-Enhanced Visual-Inertial Navigation System for Low-Speed Robots," *IEEE Trans. Instrum. Meas.*, vol. 72, pp. 1–13, 2023, doi: 10.1109/TIM.2023.3260279.

[35] J. Jiao *et al.*, "A Novel Dual-Lidar Calibration Algorithm Using Planar Surfaces," in *2019 IEEE Intelligent Vehicles Symposium (IV)*, Jun. 2019, pp. 1499–1504. doi: 10.1109/IVS.2019.8814136.

[36] X. Liu, C. Yuan, and F. Zhang, "Targetless Extrinsic Calibration of Multiple Small FoV LiDARs and Cameras Using Adaptive Voxelization," *IEEE Trans. Instrum. Meas.*, vol. 71, pp. 1–12, 2022, doi: 10.1109/TIM.2022.3176889.

[37] W. Xu, Y. Cai, D. He, J. Lin, and F. Zhang, "FAST-LIO2: Fast Direct LiDAR-Inertial Odometry," *IEEE Trans. Robot.*, pp. 1–21, 2022, doi: 10.1109/TRO.2022.3141876.

[38] S. Agarwal, K. Mierle, and The Ceres Solver Team, *Ceres Solver*. (Oct. 2023). C++. [Online]. Available: https://github.com/ceres-solver/ceres-solver

[39] T. Qin, P. Li, and S. Shen, "VINS-Mono: A Robust and Versatile Monocular Visual-Inertial State Estimator," *IEEE Trans. Robot.*, vol. 34, no. 4, pp. 1004–1020, Aug. 2018, doi: 10.1109/TRO.2018.2853729.

[40] P. Geneva, K. Eckenhoff, W. Lee, Y. Yang, and G. Huang, "OpenVINS: A Research Platform for Visual-Inertial Estimation," in *2020 IEEE International Conference on Robotics and Automation (ICRA)*, Paris, France: IEEE, May 2020, pp. 4666–4672. doi: 10.1109/ICRA40945.2020.9196524.

[41] L. von Stumberg and D. Cremers, "DM-VIO: Delayed Marginalization Visual-Inertial Odometry," *IEEE Robot. Autom. Lett.*, vol. 7, no. 2, pp. 1408–1415, Apr. 2022, doi: 10.1109/LRA.2021.3140129.

[42] K. Chen, R. Nemiroff, and B. T. Lopez, "Direct LiDAR-Inertial Odometry: Lightweight LIO with Continuous-Time Motion Correction," in *2023 IEEE International Conference on Robotics and Automation (ICRA)*, May 2023, pp. 3983–3989. doi: 10.1109/ICRA48891.2023.10160508.

[43] C. Doer and G. F. Trommer, "Radar Inertial Odometry With Online Calibration," in *2020 European Navigation Conference (ENC)*, Nov. 2020, pp. 1–10. doi: 10.23919/ENC48637.2020.9317343.

[44] Y. Xu, Q. Huang, S. Shen, and H. Yin, "Incorporating Point Uncertainty in Radar SLAM," *IEEE Robot. Autom. Lett.*, vol. 10, no. 3, pp. 2168–2175, Mar. 2025, doi: 10.1109/LRA.2025.3527344.

[45] C. Zheng *et al.*, "FAST-LIVO2: Fast, Direct LiDAR–Inertial–Visual Odometry," *IEEE Trans. Robot.*, vol. 41, pp. 326–346, 2025, doi: 10.1109/TRO.2024.3502198.

[46] J. Lin and F. Zhang, "R3LIVE: A Robust, Real-time, RGB-colored, LiDAR-Inertial-Visual tightly-coupled state Estimation and mapping package," in *2022 International Conference on Robotics and Automation (ICRA)*, 2022, pp. 10672–10678. doi: 10.1109/ICRA46639.2022.9811935.

[47] M. Grupp, *evo*. (Feb. 12, 2026). Python. [Online]. Available: https://github.com/MichaelGrupp/evo

[48] X. Niu, L. Wang, Q. Chen, H. Tang, Q. Zhang, and T. Zhang, "KF-GINS: an open-sourced software for GNSS/INS integrated navigation," *GPS Solut.*, vol. 29, no. 4, p. 202, Oct. 2025, doi: 10.1007/s10291-025-01967-w.

[49] T. Qin, S. Cao, J. Pan, and S. Shen, "A General Optimization-based Framework for Global Pose Estimation with Multiple Sensors," *ArXiv190103642 Cs*, Jan. 2019, [Online]. Available: http://arxiv.org/abs/1901.03642

[50] X. Niu, H. Tang, T. Zhang, J. Fan, and J. Liu, "IC-GVINS: A Robust, Real-Time, INS-Centric GNSS-Visual-Inertial Navigation System," *IEEE Robot. Autom. Lett.*, vol. 8, no. 1, pp. 216–223, Jan. 2023, doi: 10.1109/LRA.2022.3224367.

[51] W. Lee, P. Geneva, C. Chen, and G. Huang, "MINS: Efficient and Robust Multisensor-Aided Inertial Navigation System," *J. Field Robot.*, May 2025, doi: 10.1002/rob.22546.

[52] C. Chi, X. Zhang, J. Liu, Y. Sun, Z. Zhang, and X. Zhan, "GICI-LIB: A GNSS/INS/Camera Integrated Navigation Library," *IEEE Robot. Autom. Lett.*, vol. 8, no. 12, pp. 7970–7977, Dec. 2023, doi: 10.1109/LRA.2023.3324825.